\documentclass{article}

\usepackage[preprint]{neurips_2026}

\usepackage[utf8]{inputenc}
\usepackage[T1]{fontenc}
\usepackage{hyperref}
\usepackage{url}
\usepackage{booktabs}
\usepackage{amsfonts}
\usepackage{amsmath}
\usepackage{nicefrac}
\usepackage{microtype}
\usepackage{xcolor}
\usepackage{graphicx}
\usepackage{subcaption}
\usepackage{multirow}
\usepackage{array}
\usepackage{alphalph}
\usepackage{float}
\usepackage{booktabs}

\hypersetup{
    colorlinks=true,
    linkcolor=black,
    citecolor=blue,
    urlcolor=blue,
    filecolor=blue,
    pdfborder={0 0 0}
}

\title{
    Denoising Models Develop Human-Like \\ Perceptual Illusion Representations \\ Across Architectures}

\author{%
  Gautam Ranka\textsuperscript{\rm 1} , Paras Chopra\textsuperscript{\rm 1} \\
  \textsuperscript{\rm 1}Lossfunk\\[0.5em]
  \texttt{gautam.ranka@lossfunk.com}, \texttt{paras@lossfunk.com}
}

\begin{document}

% Add logo at the very top of the document, before maketitle
\thispagestyle{empty}
\begin{center}
\vspace*{-1cm}
\includegraphics[width=0.25\textwidth]{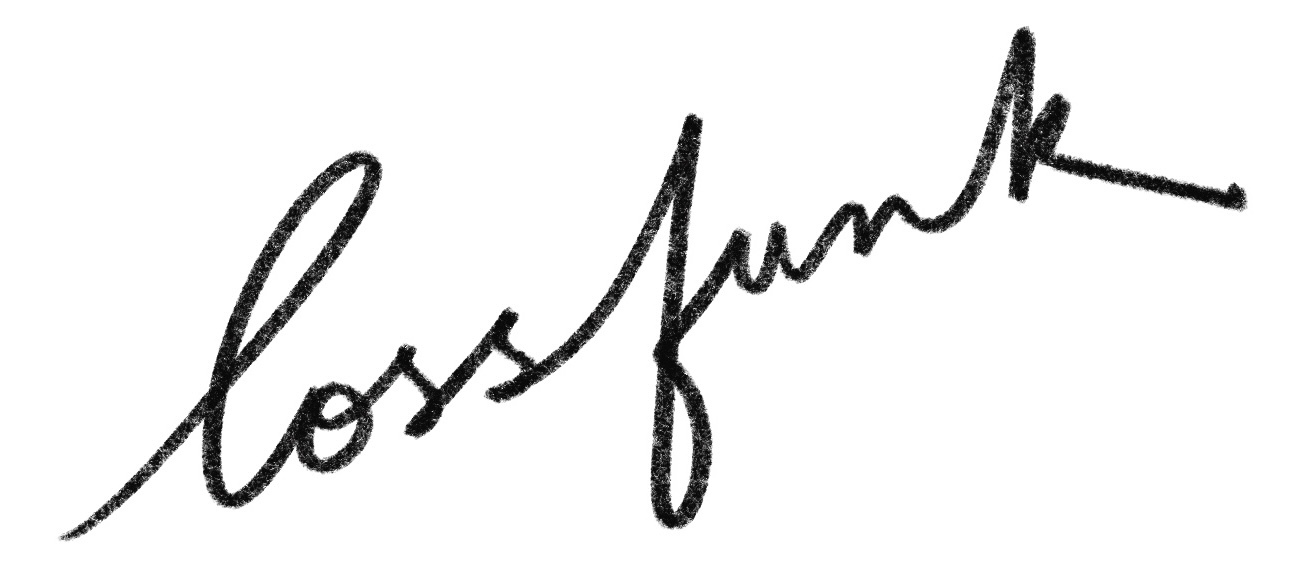}
\vspace{0.3cm}
\end{center}

\maketitle

\begin{abstract}
Deep neural networks trained on natural images are shown to produce outputs consistent with
  human observers for brightness illusions. While this phenomenon has been documented across architectures, all evidence, to date, is
measured at the output level: restored pixels, decoded trajectories, or classification
decisions. Whether these models actually \emph{represent} illusions internally, and if so
where and how, remains unknown.
We show that denoising models develop illusion-sensitive representations
at specific internal layers, across varied architectures.
Specifically, we identify the layers and channels that discriminate illusory from
physically matched control regions. We show that the denoising objective is a more important driver of the effect than the architecture. On domain-appropriate stimuli, these activations track a
validated psychophysical model of human brightness perception (FLODOG; Spearman
$\rho \geq 0.70$) and scale monotonically with parametric illusion strength.
Leveraging these findings, we provide causal evidence via channel ablation showing that illusion-sensitive channels specifically and substantially affect the \emph{internal} signal.
Yet injecting these representations into the generation pipeline produces no measurable
pixel shift across all tested architectures;
we term such representations \emph{perceptual phantoms}:  active in internal processing yet invisible to any output-based evaluation.
While related internal-output dissociations have been characterized in language models, this is the first such characterization for perceptual representations in denoising vision models.
\end{abstract}

\begin{figure}[t]
  \centering
  \includegraphics[width=\linewidth]{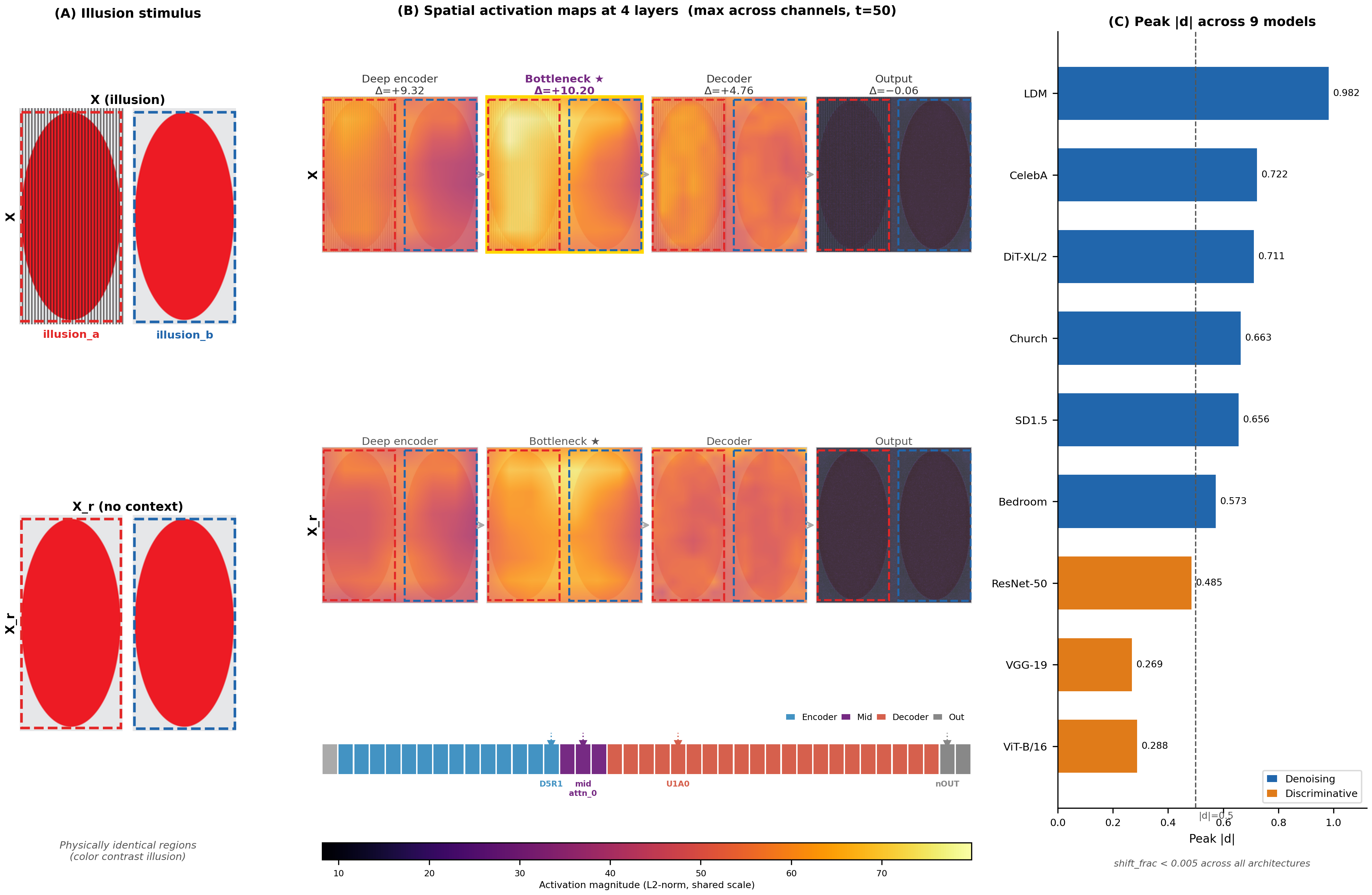}
  \caption{\textbf{Illusion encoding traced through a denoising U-Net.}
    (A)~A GVIL color contrast illusion: UP: the two elliptical regions
    (\texttt{illusion\_a}, red; \texttt{illusion\_b}, blue) are physically the
    same color but appear different due to the context. DOWN: same image with illusive feature removed
    (B)~Spatial activation maps at four
    layers tracing the signal through the network for the UP and DOWN images.  $\Delta$ denotes the raw mean activation difference between
    \texttt{illusion\_a} and \texttt{illusion\_b}.
    The illusion differential peaks at the mid-block bottleneck
    and is nearly absent at the output, a perceptual phantom.
    (C)~Peak $|d|$ (Cohen's~$d$) across nine models: every denoising model
     above the medium-effect threshold ($|d| \geq 0.5$) compared to discriminative models falling at or below.
    }
  \label{fig:hero}
\end{figure}

\section{Introduction}

Human visual illusions are not bugs in perception \-- they are probes of its computational
architecture.
The same contextual integration that makes an identically gray patch appear lighter or darker
depending on its surround underlies robust scene understanding.
Recent discoveries that deep neural networks trained on natural images spontaneously replicate
human illusion sensitivity~\citep{gomez2019convolutional,ward2019,jaini2024intriguing,gomez2024color}
suggest these biases may be a necessary consequence of learning efficient representations
of natural scene statistics~\citep{weiss2002motion}.

But does the model \emph{actually encode} the illusion, or merely produce correlated statistics?
\citep{gomez2024color} show DDIM trajectories exhibit human-like color shifts, an important
advance beyond purely output-level analysis, but still  measured in decoded pixel
intensity at intermediate timesteps, not layer-resolved internal activations. To the best of our knowledge, no published work has performed fine-grained causal interventions on illusion representations inside a denoising neural network, nor tested whether such representations generalize beyond U-Net architectures.
We close the gap between ``the model's outputs look human-like'' and
``the model's internal computations are organized like human perception.''

To answer this question we probe internal activations layer by layer across nine models:
six denoising models spanning three architecture families (pixel-space DDPMs, latent
diffusion U-Nets, and diffusion transformers) plus three discriminative baselines.
Our findings organize around three claims:
\begin{enumerate}
  \item \textbf{Denoising models encode illusions internally, regardless of architecture.}
    Every denoising model we test, including both U-Nets and diffusion transformers,
    develop illusion-sensitive representations at specific layers.
    Similar discriminative models, however, show consistently weaker effects.
    The denoising objective, not architecture, appears to be what matters.
  \item \textbf{The encoding is causally involved in internal processing.}
      Targeted channel ablation demonstrates that
      illusion-sensitive channels play a causal role in shaping the internal representations.
      Specific channels dominate the effect, and the signal concentrates at the bottleneck rather than in output-bound pathways.
  \item \textbf{Illusion encoding causally attenuates en route to the output.}
    The encoding propagates downstream, but with progressively diminishing output coupling to finally reach a point when illusion specific channels disturb pixel reconstruction less than matched random channels.  We term these representations \emph{Perceptual Phantoms} in denoising models, extending the phenomenon to generative vision.
    
\end{enumerate}

\section{Related Work}
\label{sec:related}

\paragraph{Visual illusions in neural networks.}
A growing body of work has established that DNNs produce outputs consistent with human
illusion perception.
\citep{gomez2019convolutional} trained CNNs on image restoration (denoising, deblurring) and
measured brightness and color shifts in the restored output images.
\citep{ward2019} showed VGG and ResNet classification decisions are biased by illusions
in the same direction as human perception.
\citep{gomez2022synthesis} extended illusion analysis to GANs via generated outputs, and
\citep{jaini2024intriguing} converted diffusion models into zero-shot classifiers, showing
that classification accuracy and shape bias match human levels.
\citep{bai2021predictive} showed predictive coding feedback is necessary for illusory contour perception in recurrent CNNs via coarse on/off ablation.
Theoretically, \citep{weiss2002motion} established that visual illusions arise naturally as
Bayes-optimal percepts under natural scene priors, the same priors that reconstruction objectives implicitly learn.

Most relevant to our work, \citep{gomez2024color} measured DDIM inversion trajectories at
intermediate denoising steps and found human-like color shifts \-- the first evidence that
illusion-consistent signals exist beyond the final output.
Their analysis also suggested that the training objective may be driving
the effect.
However, their metric is decoded pixel intensity at intermediate timesteps, not
layer-resolved internal activations: they show illusion effects exist \emph{along} the
latent trajectory but not \emph{where inside the model} they arise.
We show (Section~\ref{sec:results}) where the encoding resides, which channels carry it,
and whether it necessarily reaches the output.

\paragraph{Mechanistic interpretability of generative models: semantic attributes, not
perceptual representations.}
Causal tracing methods developed for language models~\citep{meng2022rome,geiger2021causal,conmy2023acdc}
have been adapted to diffusion models:
\citep{basu2023editability} performed causal mediation in text-to-image models, finding
distributed rather than localized knowledge;
\citep{kwon2023diffusion} identified the U-Net bottleneck as a semantic latent space;
\citep{hertz2022prompt} showed cross-attention causally controls spatial layout.
However, all this work targets \emph{semantic} attributes \-- object identity, style,
spatial relations.
No causal tracing of \emph{perceptual} representations (encoding \emph{how} stimuli are
perceived, not \emph{what} they depict) has been reported.

Standard causal tracing~\citep{meng2022rome,basu2023editability}
works by patching activations and measuring the \emph{output change} by construction, and is therefore structurally unable to detect representations that do not propagate to outputs (Section ~\ref{sec:causal}). We use internal-to-internal causal methods that measure intervention effects at intermediate layers.

\paragraph{Internal-output dissociation in interpretability.} The observation that internal representations can be detectable yet absent from model behavior has a substantial prior literature. \citep{hewitt-liang-2019-designing}  formalized this as a probing concern, introducing control tasks for selectivity. \citep{elazar-etal-2021-amnesic} used iterative null-space projection to remove probe-detected linguistic properties from language model representations and found several were not behaviorally used. \citep{makelov-etal-2024-subspace} formalized "harmless" (null-space) versus "pernicious" (hidden-pathway) divergences from causal interventions. \citep{fakhar-etal-2024-downstream} demonstrated that activity-causal contribution dissociation may be generic in nonlinear networks. On the methodology side, \citep{canby-etal-2024-reliability} showed nullifying interventions are systematically less complete than counterfactual ones, and \citep{zhang-nanda-2024-activation} recommend evaluating multiple intervention modalities. Our work characterizes this phenomenon for perceptual representations in denoising vision models, a domain not covered by prior literature, with cross-architecture replication and an attenuation-gradient analysis that goes beyond binary internal/output dissociation.

\section{Methods}
\label{sec:methods}

\subsection{Models and Stimuli}

\paragraph{Denoising models and Baselines.}
We study three architecture families:
(1)~\emph{Pixel-space DDPMs}: \texttt{google/ddpm-ema-church-256}~\citep{ho2020ddpm}
(113.7M parameters, 256$\times$256, LSUN Churches; primary model),
\texttt{google/ddpm-ema-bedroom-256} (LSUN Bedrooms), and
\texttt{google/ddpm-ema-celebahq-256} (CelebA-HQ faces);
(2)~\emph{Latent diffusion U-Nets}: \texttt{CompVis/ldm-celebahq-256}~\citep{rombach2022latent}
and Stable Diffusion 1.5 (latent 64$\times$64, VAE encoder/decoder);
(3)~\emph{Vision transformer}: DiT-XL/2~\citep{peebles2023scalable} (28 transformer blocks,
675M parameters, ImageNet class-conditioned, no U-Net structure). Discriminative models of different architectures are chosen as baselines to evaluate effect of training objective.~(Appendix~\ref{app:discmodels})

All models are frozen throughout with single forward pass per image per timestep.

\paragraph{Stimuli.}
We use two stimulus sets.
The \emph{GVIL dataset}~\citep{zhang2023gvil} provides $N{=}35$ base color illusion images containing 2 illusion regions, together with
matched real-counterpart images $X_r$ in which the illusory context is removed and objects
are placed on a uniform white background.
The original dataset was larger ($N=72$), but had flip augmentations which we don't include in our analysis due to potential inflation of results ~(Appendix~\ref{app:n35}). 
The \emph{Pyllusion parametric stimuli} \citep{makowski2021parametric} provide synthetic Ebbinghaus and Ponzo illusions
at 10 strength levels ($N{=}10$ images per level) for dose-response analysis~(Appendix~\ref{app:Pyllusion}).

\subsection{Activation Probing Protocol}

\paragraph{Single-step forward probing.}
For each image at timestep $t$, we construct a noisy input
$$x_t = \sqrt{\bar{\alpha}_t}\, x_0 + \sqrt{1-\bar{\alpha}_t}\, \epsilon$$
,where $\epsilon \sim \mathcal{N}(0, I)$ and perform a single forward pass through the frozen
model, capturing intermediate activations via registered forward hooks.(Appendix~\ref{app:probdetails})
Primary claims are based on $t \in \{50, 150\}$; $t > 700$ is excluded due to a
heavy-noise confound (Appendix~\ref{app:t900}).

\paragraph{Channel aggregation metrics.}
Four methods reduce the activation tensor to a spatial $(H, W)$ map:
L2-norm, mean, max, and std across channels.
The \emph{max} metric is primary for observational experiments (strongest signal-to-noise
at $N{=}35$); all four are reported in supporting tables.

\paragraph{Effect sizes.}
For each layer and timestep, $\Delta$ is computed per image.
We report all three standardized effect size metrics: Cohen's $d$ (standard paired),
Hedge's $g$ (bias corrected) , and Glass's $\Delta$
(normalised by SD). (Appendix~\ref{app:probdetails})

Thresholds: $<0.2$ negligible, $0.2$--$0.5$ small, $0.5$--$0.8$ medium, $>0.8$ large.

\paragraph{A-priori layers.}
We define three a-priori layers based on known architectural significance:
\texttt{mid\_attn\_0} (bottleneck attention), \texttt{mid\_resnet\_0} (bottleneck ResNet),
and \texttt{down\_5\_resnet\_1} (deepest encoder layer).
These were selected before observational analysis based on the architectural hypothesis that
the bottleneck concentrates illusion information; all subsequent analyses at these layers
are confirmatory, not exploratory.

\subsection{Control Experiments}

We run (i) \textbf{Pixel-Shuffle} within illusion\_a to dissociate gestalt vs. feature processing followed by (ii) \textbf{Real-Counterpart} comparison ($\Delta ill/ \Delta real$) to seperate illusion effect from image context; (iii) \textbf{Random Initialization} to separate training from architecture; (iv) \textbf{Random-Image} baselines with matched mask geometry, formally tested via d\_excess \ref{app:geometry_excess}; Finally, we do (v) \textbf{Multiple-Hypothesis correction} via 10,000-permutation tests with FDR to strengthen the claims (Appendix~\ref{app:permutation}). More details are mentioned in Appendix~\ref{app:controlexp}

\subsection{Correlational Protocol}

\paragraph{Psychophysical correlation.}
FLODOG~\citep{robinson2007flodog} is a validated computational model of human brightness
perception using oriented Difference-of-Gaussian filters with divisive normalisation, which produces pixel-level predicted brightness maps.
Spearman $\rho$ is computed between FLODOG predictions and activation magnitudes.
However, FLODOG is a luminance model and hence is applicable only to grayscale brightness stimuli (SBC, Hermann grid). (Appendix~\ref{app:flodog_full})

\paragraph{Dose-response.}
Pyllusion\citep{makowski2021parametric} provides us with illusions at different strength levels. We use it to create stimuli at 10 strength levels, all processed through the same single-step forward
probing, region comparison, and effect-size computation described earlier.
Spearman $\rho$ between illusion strength and Glass's $\Delta$ (or Cohen's $d$) quantifies the monotonicity of the dose-response relationship.

\subsection{Causal Intervention Protocol}
\label{msubsec:causal}

\paragraph{Channel ablation.}
Channels at \texttt{mid\_resnet\_0} that exceed $|d| \geq 0.5$ per the per-channel analysis are zeroed, and the percentage reduction in $\Delta$ at \texttt{mid\_attn\_0},
identified as the peak observational layer, is measured.
Separate ablations are performed for \texttt{pos\_d} ($d \geq 0.5$),
\texttt{neg\_d} ($d \leq -0.5$), and \texttt{all\_sig} ($|d| \geq 0.5$) channels.
To ensure results are not threshold-specific, we repeat the analysis
across six effect-size cutoffs ($|d| \in \{0.2, 0.3, 0.5, 0.8, 1.0, 1.2\}$).

To test whether the ablation reduction is specific to illusion-sensitive channels or reflects
generic capacity degradation, we ablate 500 random channel sets of matched count (140 channels
each) at \texttt{mid\_resnet\_0} and measure the resulting $\Delta$ reduction at
\texttt{mid\_attn\_0}, constructing a null distribution (Appendix~\ref{app:random_channel}).

To address the concern that channels are selected and evaluated on the same $N{=}35$ images,
we perform 5-fold and leave-one-out (LOO) cross-validation (Appendix~\ref{app:crossval}).

\paragraph{Skip-connection probing.}
To test whether the illusion signal propagates via U-Net skip connections, we compare $|d|$
at the bottleneck (\texttt{mid\_attn\_0}), skip-proxy layers (encoder last-resnets),
and decoder-entry layers.
Concentration at the bottleneck with low skip-proxy values indicates the signal is not
transmitted to the decoder via skip connections.

\paragraph{Phantom injection.}
(a)~DDIM 20-step injection: each $X_r$ is reconstructed via DDIM inversion
and its denoising trajectory is modified by replacing activations at specified
layers/timesteps with those from $X$.
The pixel-level shift fraction measures whether injecting illusion representations into a clean trajectory produces visible perceptual effects.

(b)~Single-step injection: $X_r$ is noised to $t{=}50$ via a single forward
diffusion step and the same activation injection is performed; the pixel shift is measured after exactly one denoising step to rule out iterative correction. 
A second variant noises $X_r$ to $t_\text{first} \approx 980$ (SNR $\approx 0.001$),
testing whether the phantom result holds even when the starting signal is almost
entirely noise.

\paragraph{Read only Test}
Injection methods test illusion-directional pixel change but not whether illusion channels are more or less output-coupled than baseline. We measure full-image reconstruction MSE under matched ablation. The illusion image X is noised to $t\_start \  \in \ \{50, 150, 300, 600, 900\}$ and reconstructed via DDIM. We zero-ablate (i) all 140 illusion-sensitive channels at mid\_attn\_0 ($|d| \geq 0.5$), (ii) the pos\_d and neg\_d subsets separately, and (iii) five matched-count random channel sets per timestep. The primary metric is full-image MSE between the ablated reconstruction and the unablated reconstruction; lower MSE under illusion-channel ablation than under random-channel ablation indicates the illusion channels are more read-only than the population baseline

% \paragraph{Within-manifold generative ablation.}
% To eliminate the alternative explanation that off-manifold ablations lead to the phantom, we test whether
% illusion-sensitive channels causally shape pixel output when operating entirely within
% the illusion image's own activation manifold.
% The illusion image $X$ is noised to $t_\text{start}{=}150$ and reconstructed via 20-step DDIM
% with channel interventions at \texttt{mid\_resnet\_0}: (a)~zero-ablation of all\_sig
% (140ch), pos\_d (56ch), and neg\_d (84ch); (b)~scaling channel activations by a multiplicative factor
% $\alpha \in \{0.0, 0.5, 1.0, 1.5, 2.0, 3.0\}$ to test dose-response.
% Three random channel sets of matched count serve as controls.
% The shift\_frac in pixel luminance differential (illusion vs.\ control region) is the
% primary metric.
% A secondary analysis measures full-image MSE, spatial localization ratio (illusion/control
% region MSE), and direction cosine toward $X_r$ (Appendix~\ref{app:gen_ablation}).

\section{Results}
\label{sec:results}

\subsection{Illusion-Sensitive Activations Are Localized}
\label{sec:obs}

\begin{figure}[t]
  \centering
  \includegraphics[width=\linewidth]{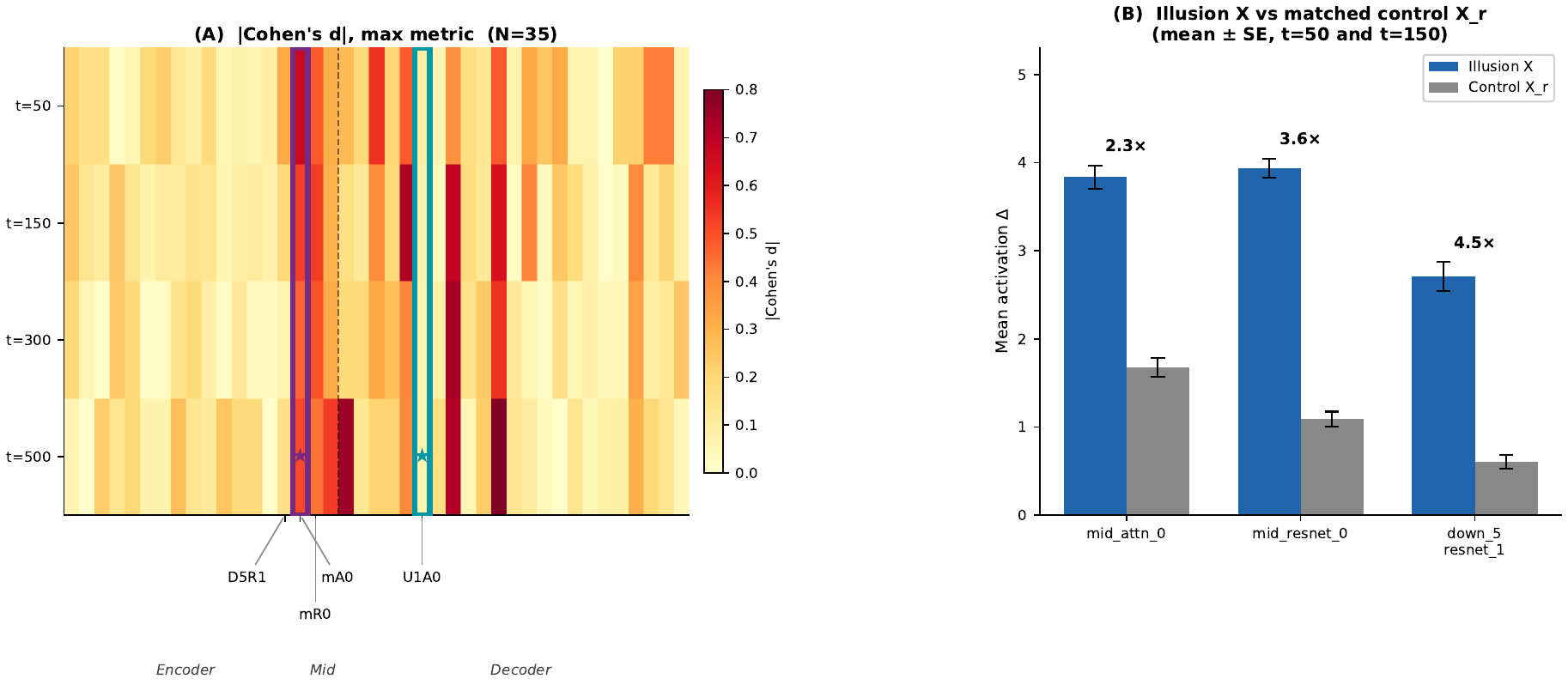}
  \caption{\textbf{Layer-resolved illusion sensitivity.}
    (A)~Heatmap of $|$Cohen's $d|$ across all 41 U-Net layers and 6 timesteps (max metric,
    $N{=}35$ base color images). \texttt{mid\_attn\_0} and
    \texttt{up\_1\_attn\_0} (largest attention effect) are highlighted.
    (B)~Paired bar chart comparing $\Delta_\text{ill}$ vs.\ $\Delta_\text{real}$ at the three
    a-priori layers, showing 2.3--4.5$\times$ specificity to illusory context.}
  \label{fig:heatmap}
\end{figure}

Figure~\ref{fig:heatmap} shows the landscape of how the effect is spread across the architecture of \texttt{google/ddpm-ema-church-256} model.
Across 41 U-Net layers and 6 timesteps, the illusion signal is not spread uniformly and concentrates sharply at the bottleneck.
The \texttt{mid\_attn\_0} layer at $t{=}50$ shows a medium-large effect
($d{=}0.663$ [max], $g{=}0.648$, Glass's $\Delta{=}{+1.392}$), consistent at $t{=}150$
($d{=}0.531$ [max], Glass's $\Delta{=}{+1.069}$).
Bootstrap resampling confirms this as robust, firmly excluding zero.
The max channel-aggregation metric is not cherry-picked: the direction is consistent across all aggregation methods at this layer (Appendix~\ref{app:n5_metrics}, \ref{app:n2_bootstrap}).

These effects are specific to illusory context, not to image region statistics:
the real-counterpart control shows the illusion image evokes 2.2--4.5$\times$ more differential activation between the two bounding-box regions than the matched image with the illusory context removed.
% (e.g., \texttt{mid\_attn\_0}: ratio $2.2\times$ at $t{=}50$; \texttt{down\_5\_resnet\_1}:
% $5.0\times$ at $t{=}150$).
The real-counterpart effect at \texttt{mid\_attn\_0} (d${=}0.330$ [L2]) is directionally consistent but underpowered at $N{=}35$ (Appendix~\ref{app:n2_bootstrap}) .

The illusion effect at mid-block layers also substantially exceeds what mask geometry alone can explain. We find that random-pixel images with identical spatial masks produce lower (often negative) $d$ at these layers; the $d_\text{excess} = d_\text{illusion} - d_\text{random}$ at \texttt{mid\_attn\_0}, $t{=}50$ is ${+1.100}$, and at \texttt{mid\_resnet\_0}:
${+1.092}$, both CIs well above zero (Appendix~\ref{app:geometry_excess}).
Early decoder layers, however, show significantly \emph{negative} $d_\text{excess}$ (random pixels $>$ illusion images), confirming that mid-block illusion sensitivity is not a mask-shape artifact.

The peak claim also survives multiple-hypothesis correction.
Permutation testing across all 246 layer$\times$timestep combinations yields
$p < 0.0001$ for the overall peak.
At the a-priori \texttt{mid\_attn\_0}, $t{=}50$: permutation $p{=}0.028$;
at \texttt{mid\_resnet\_0}, $t{=}50$: $p{=}0.017$
(Appendix~\ref{app:permutation}).

% \begin{figure}[t]
%   \centering
%   \includegraphics[width=\linewidth]{figures/fig4_pixel_shuffle}
%   \caption{\textbf{Pixel-shuffling dissociation reveals dual processing pathways.}
%     (A)~Effect size ($d$) before and after pixel shuffling of the \texttt{illusion\_a} region,
%     for the three a-priori layers.
%     The mid-block (\texttt{mid\_attn\_0}, \texttt{mid\_resnet\_0}) collapses ($d: 0.663 \to 0.206$
%     and $1.392 \to 0.076$ Glass's $\Delta$); the deep encoder (\texttt{down\_5\_resnet\_1})
%     is preserved and slightly increases at $t{=}150$.
%     (B)~Schematic of the dual-pathway architecture identified in the U-Net:
%     a gestalt-sensitive mid-block pathway and a feature-based deep-encoder pathway.}
%   \label{fig:shuffle}
% \end{figure}

We find that this localization reflects two distinct processing modes. Pixel Shuffling \emph{within} the \texttt{illusion\_a} region,
preserving their local marginal statistics but destroying spatial arrangement, leads to mid-block effect collapse 
In contrast, \texttt{down\_5\_resnet\_1} (deep encoder) is preserved after shuffling, identifying a \emph{feature-based} pathway that responds to local texture and contrast statistics.

% A scrambled-context control (Experiment~5; Appendix~\ref{app:n3_scrambled}) provides additional
% negative evidence: rotating the global surround while preserving the target bbox crops does not
% selectively reduce mid-block activation differences (all CIs include zero at mid-block layers).
% This shows the far surround outside the annotated regions is not the driver, though the bounding box
% crops retain local background pixels from the original scene, leaving the contribution of
% immediate local context unresolved.

% This dissociation is reminiscent of functional specialization in biological vision,
% where early visual areas encode local contrast and orientation while higher areas integrate
% scene-level spatial context.
% We note this as a suggestive analogy rather than a mechanistic equivalence:
% the pixel-shuffling control is too coarse to establish correspondence with specific cortical
% pathways.
% Nonetheless, the finding that diffusion U-Nets develop functionally distinct processing
% streams under the sole pressure of denoising natural images is consistent with the hypothesis
% that efficient natural scene coding creates convergent computational
% organization~\citep{olshausen1996emergence}.

\subsection{Activations Track Psychophysical Perceptions}
\label{sec:flodog}

\begin{figure}[t]
  \centering
  \includegraphics[width=0.7\linewidth]{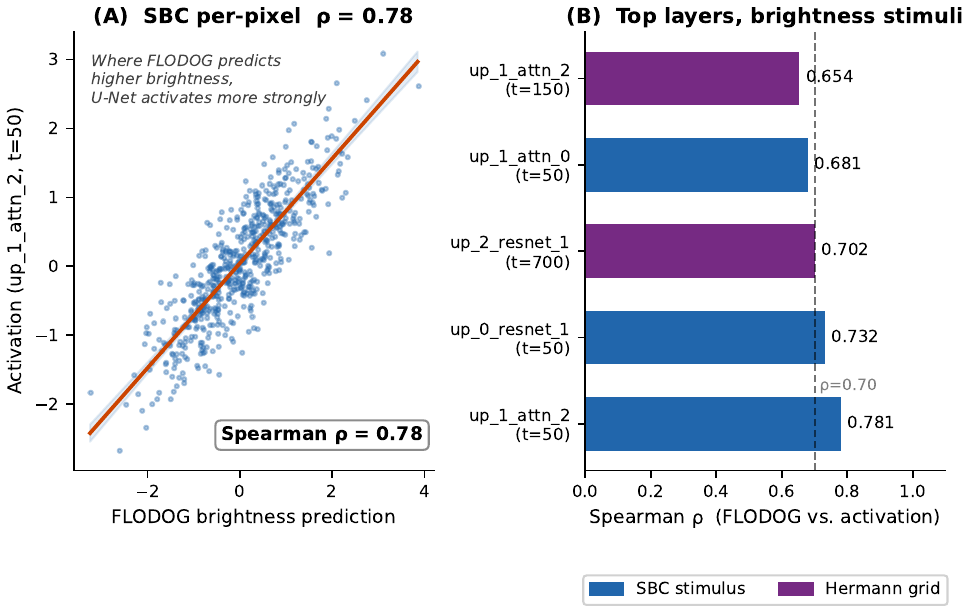}
  \caption{\textbf{Psychophysical alignment of U-Net activations.}
    (A)~Per-pixel scatter plot of FLODOG-predicted brightness vs.\ U-Net activation magnitude
    within the illusion region for a representative SBC image, at \texttt{up\_1\_attn\_2}, $t{=}50$
    ($\rho{=}0.78$).
    (B)~Top layers by Spearman $\rho$ on programmatic SBC and Hermann-grid stimuli.}
  \label{fig:flodog}
\end{figure}

FLODOG~\citep{robinson2007flodog} is a computational model of human
brightness perception, validated for luminance-based grayscale illusions (SBC, White's Effect, Hermann grid) but not for chromatic contrast effects.
On domain-appropriate programmatic stimuli, U-Net activations show strong psychophysical alignment: $\rho{=}0.78$ at \texttt{up\_1\_attn\_2}, $t{=}50$ for SBC, and $\rho{=}0.70$ at \texttt{up\_2\_resnet\_1}, $t{=}700$ for Hermann-grid stimuli.
Multiple layers at different timesteps exceed $\rho \geq 0.65$ (Figure~\ref{fig:flodog}B).
This means that regions where humans perceive stronger brightness contrast show proportionally
higher U-Net activation pointing towards a continuous, graded correspondence, not merely a binary detection.

When tested across the full $N{=}35$ GVIL color illusion dataset, which includes
chromatic contrast effects outside FLODOG's valid domain, the aggregate correlation collapses
to near zero (mean $\rho{=}{-0.019}$).
This domain mismatch is expected and informative: it suggests the U-Net separates luminance and
chromatic processing, paralleling known V1/V2 vs.\ V4 dissociations in human vision though this is currently speculative.
(see Appendix~\ref{app:flodog_full}).

\begin{figure}[t]
  \centering
  \includegraphics[width=\linewidth]{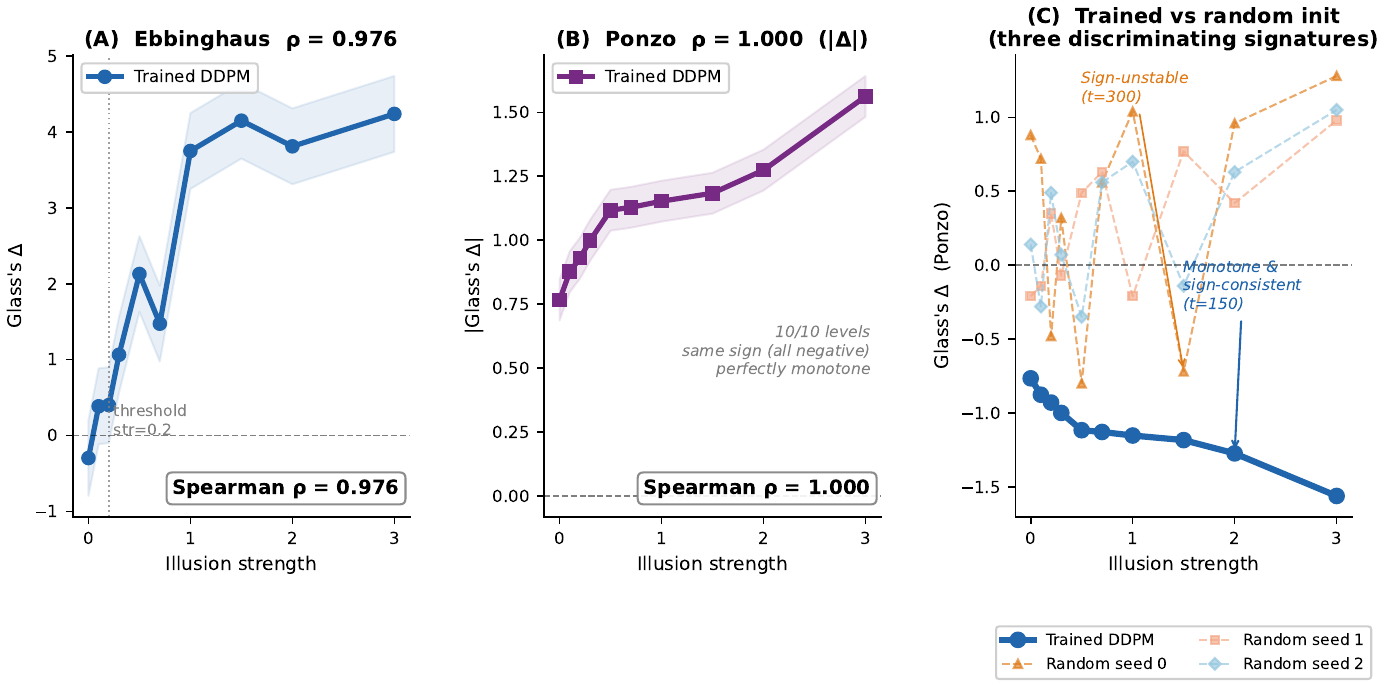}
  \caption{\textbf{Monotonic dose-response to parametric illusion strength.}
    (A)~Ebbinghaus: Glass's $\Delta$ vs.\ illusion strength at \texttt{up\_1\_attn\_2}, $t{=}50$.
    Signal is suppressed at low strengths (str$\,{<}0.2$) and rises monotonically above a threshold.
    (B)~Ponzo: perfectly monotone dose-response at \texttt{conv\_norm\_out}, $t{=}150$
    (C)~Trained vs.\ random-initialization comparison at the trained model's best layer. }
  \label{fig:doseresponse}
\end{figure}

To compare the alignment on geometric illusions as an exploratory study, we use the Dose-Response experiment which uses different strength illusions to obtain correlation between strength and Glass's $\Delta$ (Figure~\ref{fig:doseresponse}). 
Overall, we find that dose-response results show clean monotonic scaling.
For Ebbinghaus illusions, Glass's $\Delta$ at \texttt{up\_1\_attn\_2}, $t{=}50$ rises
with Spearman $\rho{=}0.976$ on 10 strength levels ($N{=}10$ per level).
For Ponzo, \texttt{conv\_norm\_out} at $t{=}150$ achieves a perfectly monotone dose-response
($\rho{=}1.000$), with Glass's $\Delta$ rising from $-0.77$ to $-1.56$.

% A cross-illusion finding: the encoder-bottleneck attention layer
% \texttt{down\_4\_attn\_0} is the top attention layer for both Ebbinghaus and Ponzo,
% consistent with the mid-block encoding identified via channel ablation.

Untrained U-Nets also achieve high $\rho \approx 0.95$ on these stimuli but with seed-inconsistent direction, confirming that the signed dose-response, not the absolute correlation, carries the training-specific signal (Figure~\ref{fig:doseresponse}C).

% Hence these three signatures discriminate trained from random models:
% (i)~timestep localization (trained peaks at $t{=}50$; random at $t{=}300$, seed-stable);
% (ii)~layer specificity (trained and random models' best layers are consistently different);
% (iii)~signed direction consistency (trained Ponzo: all 10 strength levels share the same sign;
% random Ponzo: sign oscillates).
%->already in figure

\subsection{Causal Flow and Phantom Property}
\label{sec:causal}

\begin{figure}[t]
  \centering
  \includegraphics[width=\linewidth]{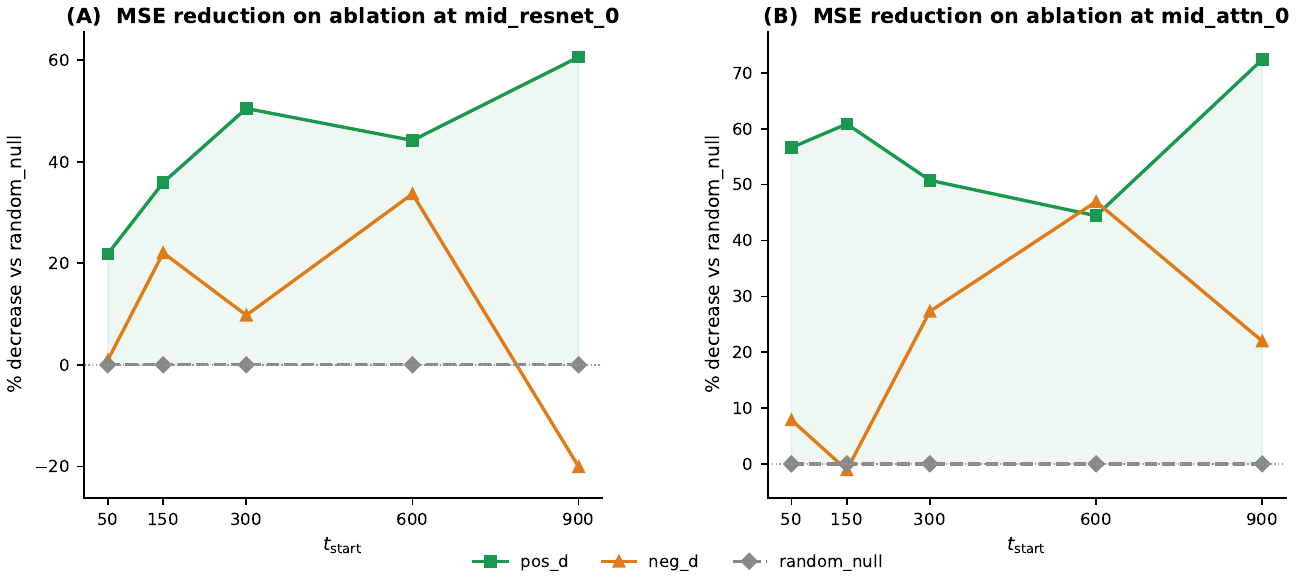}
  \caption{Percentage decrease in Reconstruction MSE on ablating specific channels vs random channels in (A) $mid\_resnet\_0$ and (B) $mid\_attn\_0$. Higher decrease means more read-only. Encoded signal becomes more read-only as signal moves downstream. }
  \label{fig:readonly}
\end{figure}

We trace the illusion signal causally through the U-Net using channel ablation. The pattern is one of progressive attenuation: the signal causally propagates from upstream layers but with diminishing effect on output, terminating in a complete phantom at the pixel level. We document this attenuation in three steps: (i) ablating illusion-sensitive channels at mid\_resnet\_0 reduces $\Delta$ at mid\_attn\_0 by 43.8\%, (ii) ablating illusion-sensitive channels at mid\_attn\_0 produces less output disturbance than random channels (the read-only asymmetry), and (iii) cross-architecture injection produces no measurable pixel shift.

\paragraph{Channel ablation.}
Per-channel analysis identifies ${\sim}27\%$ of channels (408/1,536 across three
a-priori layers) with $|d| \geq 0.5$. Zeroing all 140 significant channels at \texttt{mid\_resnet\_0} ($|d| \geq 0.5$) reduces the illusion signal at \texttt{mid\_attn\_0} by \textbf{43.8\%} (CI excludes zero) (Appendix~\ref{app:channel_ablation_detail}).
This reduction is \emph{specific} to illusion-sensitive channels:
ablating 500 random channel sets of matched count (140 channels) produces a mean reduction of only $4.7\%$ (specificity ratio 8.3×, p=0.008; Appendix~\ref{app:random_channel}). The effect survives LOO cross-validation with expected shrinkage (median $+21.8\%$, 91\% of held-out images positive; Appendix~\ref{app:crossval}). The primary driver is the neg\_d (suppressor) subset; pos\_d alone produces near-null internal effects, foreshadowing their phantom signature at output.

\paragraph{Read-only signature at the output-coupled layer.}
While ablation at mid\_resnet\_0 propagates downstream to mid\_attn\_0, ablation at mid\_attn\_0 -- the peak observational layer and the closest U-Net layer to the decoder -- reveals the read-only signature. We measure full-image reconstruction MSE under matched ablation across multiple start timesteps for the 140 illusion-sensitive channels, the pos\_d and neg\_d subsets, and five matched-count random channel sets per timestep. At the primary timesteps $t   \in \{50, 150\}$, illusion channels produce 9–14\% lower MSE than random channels, with the pos\_d subset alone 57–61\% lower. Crucially, the same pos\_d channels measured one layer upstream at mid\_resnet\_0 are only 22–36\% read-only (Figure~\ref{fig:readonly},  Appendix~\ref{app:phantommse}). The read-only effect strengthens as the signal flows downstream. 

\paragraph{Phantom at output.}
To understand the significance of the MSE signature, we test these via the Phantom Injection tests by measuring the shift fractions. $$ shift\_frac = (\Delta\_patched - \Delta\_clean) / (\Delta\_X - \Delta\_Xr \ + \epsilon )$$
Shift fraction measures the pixel transfer ratio for activations injected into a clean reconstruction trajectory. A value of 0 means zero pixel-level effect; a value of 1 would mean complete transfer of the illusion's pixel signature. Across four architectures (Church DDPM, Bedroom DDPM, LDM-CelebA-HQ, DiT-XL/2), $shift\_frac \approx 0$ with all CIs including zero (Appendix~\ref{app:n1_phantom}, \ref{app:phantomarch}, ~\ref{app:permodelpeak}). We also rule out iterative correction via single step injections and off-manifold injections via separate within-manifold ablation (Appendix~\ref{app:gen_ablation}). The perceptual phantom is universal. 

\paragraph{Robustness across intervention modalities.}
Recent work has shown that nullifying interventions (e.g., zero-ablation) can be systematically incomplete: ablated information may remain recoverable via residual pathways, producing spurious null results \citep{canby-etal-2024-reliability}. We rule this out by replicating the phantom under counterfactual interventions: cross-image activation patching mentioned above, and multiplicative amplification up to 3× (Appendix~\ref{app:positive_injection}) both produce no measurable pixel shift. The MSE asymmetry runs opposite to the direction predicted by incomplete nullification i.e. the illusion channels disturb output \textit{less} than random.

\paragraph{Mechanism}
We then try to analyze the effect mechanistically. For DiT, we trace injected activations block by block. Injection at block~26 produces a massive perturbation at block~27, but self-attention and LayerNorm within that single block absorb it entirely (Appendix~\ref{app:dit_mechanism}).
% Even saturating six consecutive blocks (22--27) with the illusion image's activations produces mean pixel shift~$\approx 0$ with random directionality.
The transformer's self-correcting dynamics provide a  mechanistic
explanation for the DiT phantom. For diffusion U-Nets, the skip-connection dissociation (Appendix~\ref{app:skip} ,\ref{app:channel_ablation_detail})
shows the illusion signal concentrates at the bottleneck but is \emph{not} transmitted via skip connections to the decoder and stays localized.

We term such representations as \emph{Perceptual Phantoms} found in denoising models: internally encoded perceptual properties that are involved in
internal processing but progressively attenuate to zero before reaching the generated image. To our knowledge, this is the first characterization of the phenomenon for perceptual representations in generative vision models, and the first to document the attenuation gradient through cross-layer causal analysis.

\subsection{Architecture Generalization and Effect of Training Objective}
\label{sec:archgen}

\begin{figure}[t]
  \centering
  \includegraphics[width=\linewidth]{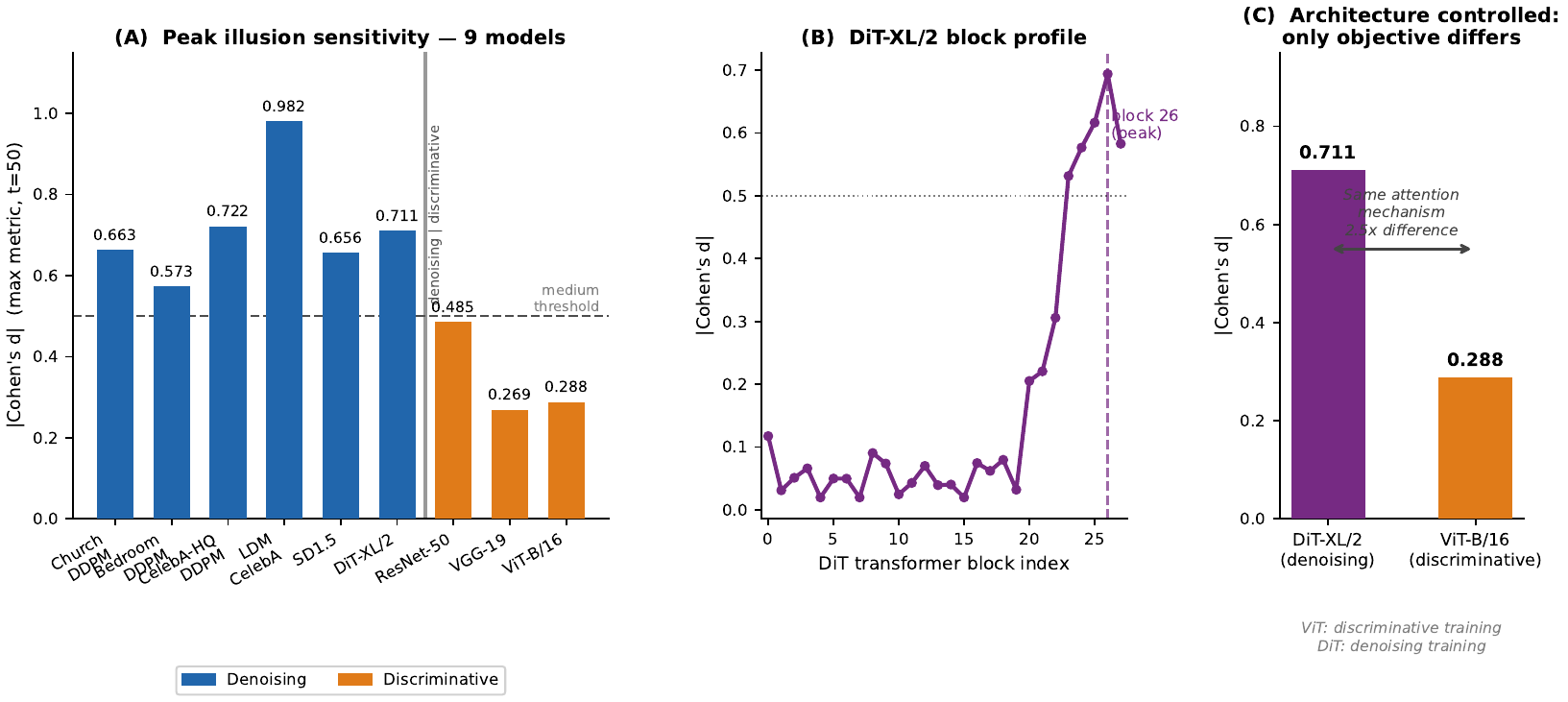}
  \caption{\textbf{Architecture generalization across denoising and discriminative models.}
    (A)~Peak $|d|$ bar chart across all nine models, colored by family
    (denoising vs.\ discriminative).
    (B)~DiT-XL/2 block profile: $|d|$ across 28 transformer blocks, peaking at block~26.
    (C)~ViT-B/16 vs.\ DiT-XL/2,
    $2.5\times$ difference in illusion sensitivity.}
  \label{fig:crossarch}
\end{figure}

All U-Net denoising models develop medium-to-large illusion sensitivity compared to the discriminative models being less than the medium threshold. However the locus in denoising models vary with training domain and architecture (Figure~\ref{fig:crossarch}A, Appendix~\ref{app:permodelpeak}). To solve the potential confound of asymmetric input, we also probe clean and noisy inputs on denoising and discriminatory models respectively finding the discriminative models to peak below the DDPM clean baseline (Appendix~\ref{app:noise_matched}).

% \begin{center}
% \begin{tabular}{llcc}
% \toprule
% Model & Peak Layer & Peak $|d|$ & Architecture \\
% \midrule
% Church DDPM (primary) & \texttt{mid\_attn\_0} & 0.663 & Pixel-space \\
% Bedroom DDPM & \texttt{up\_5\_resnet\_1} & 0.573 & Pixel-space \\
% CelebA-HQ DDPM & \texttt{up\_0\_resnet\_1} & 0.722 & Pixel-space \\
% LDM-CelebA-HQ & \texttt{up\_0\_attn\_2} & \textbf{0.982} & Latent \\
% SD1.5 & \texttt{sd15\_mid\_attentions.0\_sa} & 0.656 & Latent \\
% \midrule
% ResNet-50 & \texttt{layer4\_block0} & 0.485 & Discriminative \\
% VGG-19 & \texttt{vgg\_pool1} & 0.269 & Discriminative \\
% ViT-B/16 & \texttt{vit\_block\_08} & 0.288 & Discriminative \\
% \bottomrule
% \end{tabular}
% \end{center}

DiT-XL/2 shows max $|d| = 0.711$ at block~26, $t{=}50$, with 11/28 blocks exceeding $|d| \geq 0.5$. ViT-B/16 and ViT-L/16 also uses the \emph{same} patch-based self-attention as DiT-XL/2, but peaks at $|d|{=}0.288$ and $|d|{=}0.207$ respectively, decoupling the effect from number of parameters (Figure~\ref{fig:crossarch}C). 
The architectural mechanism is shared between them and what differs is the objective
(classification vs.\ denoising), making this a clean evidence that the denoising objective drives the effect heavily.

Finally, to isolate the effect of architecture from training, we compare the  models with their randomly initialized seeds. For Church-DDPM, \texttt{up\_1\_attn\_0} layer shows higher $d = +0.952$ than random seeds with \emph{opposite} signs, pointing that training reverses the architectural default. Trained DiT shows peak $|d|{=}0.711$, higher than the median $d {=}0.364$. However seed 0 ($|d|=0.741$) approximates the trained effect(Appendix~\ref{app:generalarchvtrain}, \ref{app:dit_multiseed}).

We also try to investigate the source of the phantom effect in DiT.
We do a block-by-block tracking after injection at block~26,
At block~27 (first to see injected activations), the injection creates a massive perturbation.
However, the self-attention and LayerNorm within block~27 absorb it the activation distribution is restored to near-native, and the signal does not survive to the output. Even saturating six consecutive blocks cannot overcome the network's self-correcting
dynamics.

Across all comparisons, the denoising objective is the common thread. Architecture and training on natural images may shape where the effect manifests, but the denoising reconstruction task makes it large.

\section{Discussion}
\label{sec:discussion}

\citep{gomez2024color} report that DDIM inversion trajectories exhibit human-like brightness and color shifts at intermediate denoising steps, attributing this to diffusion models mapping illusory stimuli onto the learned natural image manifold. This appears in tension to our findings of internal computations that never reach outputs. We believe their trajectory-level shifts likely reflect frequency-dependent noise destroying context faster than target. Together, the two findings suggest that diffusion models process illusion-relevant information at multiple levels (trajectory-level and representation-level).

Overall, our results show that training objective is a much higher contributor to human-like illusion sensitivity, than the architecture. We show this by using multiple architectures and especially the ViT-B/16 and ViT-L/16 vs.\ DiT-XL/2 gap, which share a similar attention based mechanism. FLODOG alignment on luminance stimuli, monotonic dose-response and mask-geometry baselines together rule out other trivial explanations. 

% We find the existence of phantoms within the denoising models. A representation can be causally involved in intermediate processing yet inert at the output which standard output-based causal tracing cannot detect by construction. This \emph{Perceptual phantom} finding is strong, replicating across multiple denoising model architectures.
% survives within-manifold testing, and
% has a mechanistic explanation in both U-Nets and DiT.

Our findings extend a growing literature on internal-output dissociation in neural networks. Prior work has characterized this for linguistic properties \citep{elazar-etal-2021-amnesic} and as a  generic activity-causal phenomenon in nonlinear networks \citep{fakhar-etal-2024-downstream}. We show the first characterization of this phenomenon for perceptual representations in generative vision models via cross-architecture replication across different architectures and an attenuation-gradient analysis showing that the dissociation is not binary.

% The FLODOG correlation and perceptual similarity on visual illusions drives us to believe that denoising-based reconstruction objective over natural images makes the models learn representations similar to human perception, aligning highly and strengthening \citep{jaini2024intriguing}'s observations. Perhaps the most intriguing question that the phantom concept raises is whether biological
% vision harbors analogous ``silent'' representations i.e. perceptual computations that
% the brain performs but never uses to guide behavior. 
\paragraph{Broader Impacts.} Denoising training appears to produce richer internal perceptual representations than discriminative training on matched architectures for some specific cases, while simultaneously attenuating a subset of these representations before output. Generative vision models may therefore encode features that are active in internal processing but undetectable from outputs alone. This internal-output asymmetry is a positive contribution toward more rigorous interpretability, but also a caution: behavioral evaluations may systematically miss what these models have actually learned to encode. 

\vspace{-0.5em}
\paragraph{Limitations and future work.}
\begin{enumerate}
  \item \textbf{Small $N$.}
    $N{=}35$ is the primary limitation.
    While bootstrap CIs confirm headline claims, some secondary results do remain underpowered. We avoided using synthetic and ablated images to avoid inflated results like the 89\% increase in measured ablation reduction. (Appendix~\ref{app:n35}).

  % \item \textbf{Causal evidence is moderate-to-strong, not definitive.}
  %   Channel ablation gives ${\sim}44\%$ reduction (in-sample; ${\sim}22\%$ cross-validated),
  %   confirmed specific to illusion channels. The illusion signal is distributed across hundreds of channels and do not provide a clean causal circuit. 
  %   Future work using sparse intervention methods (e.g.,
  %   ACDC~\citep{conmy2023acdc}) at larger $N$ could identify a minimal sufficient channel set.

  \item \textbf{Psychophysical alignment limited to luminance.}
    FLODOG $\rho $ applies only to SBC/Hermann. Developing or adapting chromatic models would enable graded psychophysical validation on the full GVIL stimulus set.

  \item \textbf{Training-vs-architecture is complex.}
    Trained models consistently exceed random median, but some random seeds match trained
    performance in DiT and dose response ($\rho \approx 0.95$, sign inconsistent)
    The denoising objective lies as the common thread, however individual-seed variability means we
    cannot fully separate effect of architecture from training.

  \item \textbf{Phantom characterization is empirical, not exhaustive.}
    We establish the phantom property for one specific class of representations specifically. We do not claim that all perceptual representations in denoising models are phantoms, nor that the phenomenon is unique to perception. Characterizing which representations are read-only versus read-write remains open.

  % \item \textbf{Dose-response architectural confound.}
  %   Albeit with sign inconsistency, random models do achieve $\rho \approx 0.95$ on parametric geometric stimuli.
  %   Hence we present this evidence as supplementary.
\end{enumerate}

\section{Acknowledgements}

We thank the JarvisLabs Team for providing the compute resources that made this project possible, as well as AWS team for providing claude resources that helped in brainstorming and code implementation.

\section{Code}

Relevant reproduction scripts and data are provided at the GitHub Repository: \url{https://github.com/Lossfunk/Denoising-models-illusion-representations}.

% \section{Conclusion}

% Denoising neural networks including pixel-space DDPMs, latent diffusion models and 
% diffusion transformers, internally encode human-like perceptual biases toward visual
% illusions showing high correlation with psychophysical brightness predictions and monotonic dose-response to illusion strength.
% These representations are found to be causally involved in feature processing with illusion-specific
% channel ablation reducing the signal by $8\times$ more than random-channel ablation
%  but never propagating to generated outputs
% across any tested architecture even when while accounting for self correction and off manifold injection.
% The ViT-B/16 vs.\ DiT-XL/2 comparison specifically isolates the denoising objective as the critical
% factor producing $2.5\times$ stronger illusion sensitivity
% under denoising than classification training, a gap that persists when input noise
% conditions are matched.
% These \emph{perceptual phantoms} suggest that denoising natural images induces perceptual
% processing that is accessible to mechanistic interpretability but invisible in model behavior and outputs. 
% This raises the question of whether there are more such phantom representations in these models, and whether biological vision harbors analogous
% silent computations that the brain performs but never uses to guide behavior.

\bibliographystyle{plainnat}
\bibliography{references}

\appendix
\renewcommand{\thesection}{\AlphAlph{\value{section}}}

\section{Dataset: N=35 Base Images}
\label{app:n35}
The GVIL dataset~\citep{zhang2023gvil} contains 72 color images: 35 base images, 35 horizontally
flipped versions, and 2 additional images.
All analyses use only the 35 base images (filter: \texttt{'flip' not in name and not name.endswith('\_r')}).
Flipped versions are excluded to prevent near-duplicate image pairs from inflating effect size estimates. When N=72 is used,  for channel ablation,
  there is an 89\% increase in reduction of downstream signal from channel ablation, when using FDR  significance as the threshold. This shows how augmented and
  similar illusion images can inflate the signal unnecessarily. This is why we chose to use N=35 images constantly within our experiments, rather than increase the data with synthetic images and datasets.

\paragraph{Region definitions.}
Each GVIL image has annotated bounding boxes: \texttt{illusion\_a} and \texttt{illusion\_b},
two physically identical regions that appear perceptually different due to their contrasting
surrounds. The within-image difference
$\Delta = \text{mean\_act}(\texttt{illusion\_a}) - \text{mean\_act}(\texttt{illusion\_b})$
measures the total illusion-driven activation differential.
Bounding boxes are eroded 10\% inward to reduce background gradient contamination from heavy
noise timesteps.

\section{Heavy Noise Confound}
\label{app:t900}
At $t{=}900$ (90\% noise fraction), low-frequency background gradients leak into small
bounding box regions, creating spurious activation differentials independent of illusion content.
Evidence: bbox erosion reduces but cannot eliminate the effect; sign flips occur above t=700 and the magnitude scales with noise level.
All primary GVIL claims are reported mostly on $t \in \{50, 150\}$.
Domain-specific analyses on programmatic stimuli with larger, uniform backgrounds are reported
at all timesteps, as the bounding-box confound is less severe for these stimuli.

\section{Discriminative Models}
\label{app:discmodels}

ResNet-50~\citep{he2016deep} and VGG-19~\citep{simonyan2015very} (ImageNet-pretrained CNNs),
and ViT-B/16~\citep{dosovitskiy2021image} (ImageNet-pretrained vision transformer).
ViT-B/16 uses the same patch-based self-attention mechanism as DiT-XL/2 but is trained
with a classification objective, enabling a controlled comparison of training objective
with matched architecture.

\section{Probing Methodology details}
\label{app:probdetails}

For U-Nets, we capture all 41 ResNet and Attention blocks.
For DiT-XL/2, activations at each of the 28 transformer blocks are captured as
$(B, 256, 1152)$ tensors, reshaped to $16{\times}16$ spatial maps and L2-normalized.
For ViT-B/16, the CLS token is stripped; 196 patches are reshaped to $14{\times}14$
spatial maps and L2-normalized.

The primary observational metric is $|d|$ (unsigned, for detecting effects regardless
of direction); causal experiments use signed $d$ (direction matters).
Glass's $\Delta$ is used for cross-experiment comparisons where different baselines make
Cohen's $d$ non-comparable.
Because different aggregation metrics capture different aspects of the signal, the same layer
may show different $d$-values depending on the channel aggregation
  metric (L2-norm, mean, max, or std) used; we specify the metric for each
reported value.

\section{Control Experiments}
\label{app:controlexp}

\paragraph{Pixel-shuffle control.}
Pixels \emph{within \texttt{illusion\_a} } are randomly permuted, preserving marginal
color/brightness statistics while destroying spatial arrangement.
A large drop in effect size after shuffling indicates gestalt-dependence;
survival indicates feature-based processing.

\paragraph{Real-counterpart control.}
For each illusion image $X$, its matched real counterpart $X_r$ ( same image with illusion inducing components removed)  undergoes the same probing.
We compare $\Delta_\text{ill}$ (from $X$) to $\Delta_\text{real}$ (from $X_r$ at the same
bounding-box positions); the ratio $\Delta_\text{ill} / \Delta_\text{real}$ quantifies
specificity to illusory context.

\paragraph{Random initialization controls.}
The complete probing pipeline is repeated with three randomly initialized seeds of the same
U-Net architecture.
Both activation and attention experiments are compared: seed-stable differences from trained
models establish training-specific effects.

\paragraph{Random image baseline and geometry excess.}
Images with uniform random pixel content (same masks) are processed through all four
model initializations to quantify the spatial bias introduced by mask geometry alone.
To formally test whether the illusion effect exceeds this geometry-driven baseline, we
compute $d_\text{excess} = d_\text{illusion} - d_\text{random}$ at each layer and timestep,
with 95\% CIs via 10,000 bootstrap resamples (Appendix~\ref{app:geometry_excess}).

\paragraph{Multiple-hypothesis correction.}
To address the concern that peak claims are selected from 246 layer$\times$timestep combinations,
we perform 10,000 permutation tests shuffling illusion\_a/illusion\_b labels and compute
max-$|d|$ across all combinations under the null.
FDR correction (Benjamini-Hochberg) is applied across all 246 $p$-values
(Appendix~\ref{app:permutation}).

\paragraph{Noise-matched discriminative probing.}
To control for the input asymmetry between denoising and discriminative models (noisy $x_t$
vs.\ clean $x_0$), we probe the DDPM on clean images ($t{=}0$, no noise) and discriminative
models (ViT-B/16, ResNet-50, VGG-19) on noise-corrupted inputs at matched timesteps
$t \in \{50, 150, 300, 500\}$ (Appendix~\ref{app:noise_matched}).

\paragraph{Scrambled-context control.}
For each image $X$, a surrogate $X_\text{sc}$ is created by rotating the full image 180$^\circ$
(\texttt{np.rot90(img, 2)}) and pasting back the original bounding-box crops from
\texttt{illusion\_a} and \texttt{illusion\_b} at their original positions.
This preserves the local target-object appearance while disrupting the far surround context.
Since the boundary boxes contain local background pixels which are generally the illusion drivers, this control does not isolate the contribution of immediate local context, 
but it tests whether the far surround is the cause of the  observed effects.
The metric $\Delta_\text{ill}(X) - \Delta_\text{ill}(X_\text{sc})$ tests whether surround
disruption selectively reduces mid-block activation differences.

\section{Full Layer × Timestep Tables}
\label{app:tables}
Full tables of Cohen's $d$, Hedge's $g$, and Glass's $\Delta$ for all 246 layer$\times$timestep
combinations across all main experiments are provided in the supplementary data files
(available with code release).

\section{Per-Channel Rankings}
\label{app:channels}
Per-channel analysis across the three a-priori layers (512 channels each):
${\sim}27\%$ of channels (408/1,536) show $|d| \geq 0.5$; ${\sim}8\%$ (123/1,536) show $|d| \geq 0.8$.
Channel 311 achieves the largest single-channel effect across all three layers simultaneously
(\texttt{mid\_resnet\_0}: $d{=}1.636$; \texttt{down\_5\_resnet\_1}: $d{=}1.619$;
\texttt{mid\_attn\_0}: $d{=}1.590$), suggesting it is the dominant illusion-sensitive feature.

\section{SD1.5 Full Results}
\label{app:sd15}
SD1.5 mask alignment was verified: 256$\to$512$\to$64 (VAE encode) and 256$\to$64 (direct
downsampling) produce identical spatial positions (100\% pixel agreement).
Three images (10\_01, 10\_02, 13\_02) have empty masks at 64$\times$64 and are excluded.
At 8$\times$8 resolution (mid-block), 21/50 images have empty masks; results at this scale
are interpreted with caution.
Full result tables for sd15/01, sd15/02, and sd15/04 are provided in supplementary data.

\section{CelebA-HQ Cross-Domain Results}
\label{app:celeba}
The \texttt{google/ddpm-ema-celebahq-256} model (trained on face photographs) shows
comparable peak effect sizes to the church-trained model at $N{=}35$
(Mean metric, \texttt{up\_0\_resnet\_1} $t{=}150$: $d{=}0.722$, $g{=}0.706$,
Glass's $\Delta{=}0.803$), confirming the perceptual bias is not domain-specific.
Importantly, the church-specific a-priori layers (\texttt{mid\_attn\_0}, \texttt{mid\_resnet\_0})
show negligible effects in CelebA-HQ \-- the bias is present but re-encoded to different layers
when the training distribution changes, suggesting that training distribution shapes the
specific locus of encoding while the overall perceptual bias is conserved.

\section{Random Initialization Full Comparison}
\label{app:random}
Three random seeds (PyTorch default initialization) were tested across all main experimental
conditions.
The seed-stable findings are:
(1) DDPM color attention at \texttt{up\_1\_attn\_0} $t{=}50$: trained positive ($+0.952$),
all 3 random seeds negative ($-0.305$, $-0.678$, $-0.167$) \-- confirmed sign flip;
(2) DDPM geometric effects peak at $t{=}300$ for random models vs.\ $t{=}50/150$ for trained;
(3) SD1.5 color attention: random seeds' best effects are at $t{=}300$ (negative),
trained is at $t{=}50$ (positive).
The seed-unstable finding is DDPM activation at \texttt{mid\_attn\_0} $t{=}50$ (random seed 2
exceeds threshold), which should not be used as a primary training discriminator.

\section{Random Image Baseline}
\label{app:random_images}
% Random-pixel images with the same mask geometry show max$|d| \approx 1$--$4$ from mask spatial
% asymmetry alone.
At the primary result layer (\texttt{up\_1\_attn\_0} $t{=}50$):
trained model on random images: $d{=}{+0.231}$ (vs.\ $+0.952$ on real illusions, $4\times$ larger);
random seeds on random images: $d{\approx}-0.15$ to $-0.21$ (vs.\ $-0.17$ to $-0.68$ on
real illusions).
Hence, The sign of the effect is determined by the model (trained vs.\ random), not by mask geometry.

\section{FLODOG Analysis and Domain Mismatch}
\label{app:flodog_full}
FLODOG applies oriented Difference-of-Gaussians filters at 6 orientations and 5 spatial scales
($\sigma_\text{center}$ capped at 32px) followed by divisive normalization to grayscale
luminance.
It is validated for luminance-based illusions (SBC, White's Effect, Hermann grid, Mach bands)
and is explicitly \emph{not} a model of chromatic contrast.
The 35 GVIL color illusion images include simultaneous color contrast effects (colored backgrounds,
chromatic surrounds) where luminance is not the primary illusion driver.
Future work could use chromatic psychophysical models (e.g., opponent-color divisive normalisation)
as ground truth for color contrast illusions.

The per-image FLODOG correlation (Appendix~\ref{app:n2_bootstrap}) reports $N{=}26$
rather than 35 because 9 GVIL images are pure chromatic contrast stimuli with negligible
luminance variation; FLODOG produces near-uniform brightness maps for these images,
making the per-pixel Spearman $\rho$ undefined or degenerate. These 9 images are excluded
from the per-image FLODOG analysis but included in all other experiments.

FDR correction (Benjamini-Hochberg) across 246 layer$\times$timestep combinations on the N=35
aggregate: a-priori layers (\texttt{mid\_attn\_0}, \texttt{mid\_resnet\_0}) do not survive
at $\alpha{=}0.05$.
Decoder ResNet layers show significant \emph{negative} correlation ($\rho \approx -0.29$),
likely reflecting chromatic (hue) encoding in the decoder reconstruction pathway.

Importantly, activation magnitude and attention routing are mechanistically separable:
attention maps correlate near-zero with FLODOG (mean $\rho{=}{-0.06}$ globally, max
$\rho{=}0.25$ at \texttt{up\_1\_attn\_2}), while activations reach $\rho{=}0.70$.
The two subsystems encode different aspects of the perceptual computation.

\section{Channel Amplification Null}
\label{app:e20}
A variant experiment tested whether amplifying the top illusion-sensitive channels
(rather than ablating them) strengthens the pixel-level illusion effect in generated images.
Result: $\approx{} null$ . Channel amplification does not produce measurable changes in generated pixel
statistics, consistent with the phantom finding.

\section{Pyllusion Stimulus Generation}
\label{app:Pyllusion}
Ebbinghaus and Ponzo stimuli were generated using the Pyllusion library
at 10 strength levels: $[0.0, 0.1, 0.2, 0.3, 0.5, 0.7, 1.0, 1.5, 2.0, 3.0]$
with $N{=}10$ images per level (random flanker positions within each level).
All stimuli are 256$\times$256 RGB, normalized to $[-1, 1]$ to match model input conventions.

\section{Per Model Peak and Locus}
\label{app:permodelpeak}

\begin{center}
\begin{tabular}{llcc}
\toprule
Model & Peak Layer & Peak $|d|$ & Architecture \\
\midrule
Church DDPM (primary) & \texttt{mid\_attn\_0} & 0.663 & Pixel-space \\
Bedroom DDPM & \texttt{up\_5\_resnet\_1} & 0.573 & Pixel-space \\
CelebA-HQ DDPM & \texttt{up\_0\_resnet\_1} & 0.722 & Pixel-space \\
LDM-CelebA-HQ & \texttt{up\_0\_attn\_2} & \textbf{0.982} & Latent \\
SD1.5 & \texttt{sd15\_mid\_attentions.0\_sa} & 0.656 & Latent \\
\midrule
ResNet-50 & \texttt{layer4\_block0} & 0.485 & Discriminative \\
VGG-19 & \texttt{vgg\_pool1} & 0.269 & Discriminative \\
ViT-B/16 & \texttt{vit\_block\_08} & 0.288 & Discriminative \\
ViT-L/16 & \texttt{vit\_block\_00} & 0.207 & Discriminative \\

\bottomrule
\end{tabular}
\end{center}

\section{Channel Ablation}
\label{app:channel_ablation_detail}

\begin{figure}[t]
  \centering
  \includegraphics[width=\linewidth]{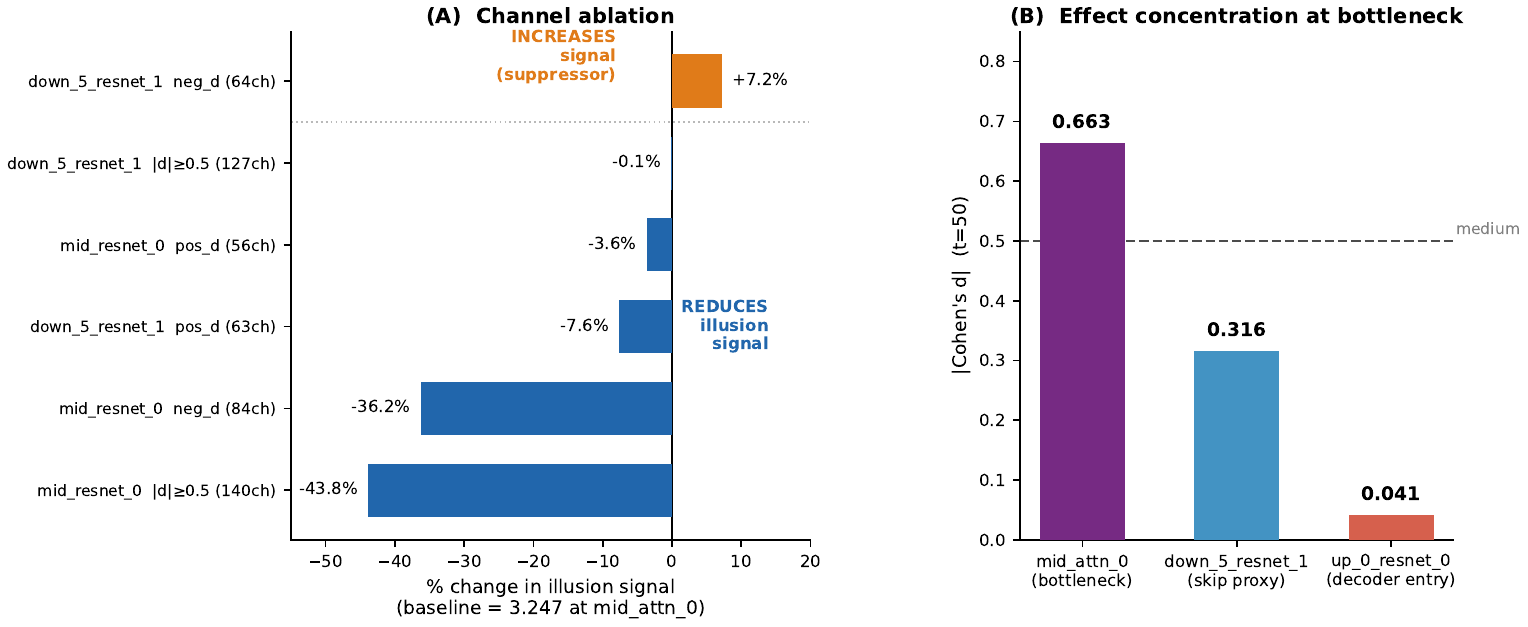}
  \caption{\textbf{Causal evidence.}
    (A)~\emph{Channel ablation}: percentage $\Delta$ reduction at \texttt{mid\_attn\_0}
    per ablation group. 
    (B)~\emph{Skip-connection dissociation}: $|d|$ comparison across bottleneck,
    skip-proxy layers, and decoder-entry layers.
    The illusion signal concentrates at the bottleneck, not in skip connections.}
  \label{fig:causal}
\end{figure}

Per-channel analysis identifies ${\sim}27\%$ of channels (408/1,536 across three
a-priori layers) with $|d| \geq 0.5$, with the most illusion-sensitive individual channel(ch.~311, \texttt{mid\_resnet\_0}) reaching $d{=}1.636$, $g{=}1.600$,
Glass's $\Delta{=}1.790$, nearly 3$\times$ the layer-level effect.
Zeroing all 140 significant channels at \texttt{mid\_resnet\_0} ($|d| \geq 0.5$) reduces the
illusion signal at \texttt{mid\_attn\_0} by \textbf{43.8\%} (bootstrap 95\% CI:
$[23.4\%, 180.2\%]$; CI excludes zero; Cohen's $d{=}0.849$, CI $[{+0.553}, {+1.239}]$)~(Figure~\ref{fig:causal}A).
The upper CI bound exceeding 100\% indicates that in some bootstrap samples, ablation
reverses the sign of the illusion signal, consistent with the \texttt{neg\_d} channels' inhibitory role.

This reduction is \emph{specific} to illusion-sensitive channels.
Ablating 500 random channel sets of matched count (140 channels) produces a mean reduction
of only $4.7\%$ (null 95\% CI: $[{-21.4\%}, {+28.2\%}]$) (Appendix~\ref{app:random_channel}).

The effect survives cross-validation with expected shrinkage
(Appendix~\ref{app:crossval}).
Leave-one-out CV yields a median reduction of $\mathbf{{+}21.8\%}$, with $91\%$ of held-out images (32/35) showing positive
reduction, confirming the  downstream causal claim survives selection bias with a medium effect.
All 140 channels are maximally stable across all CV splits.

A threshold sensitivity analysis sweeping shows the
effect is not threshold-specific with both the \texttt{neg\_d} and \texttt{all\_sig} groups produce reliable
positive reductions across all thresholds 0.2--0.8, with CIs excluding zero
(Appendix~\ref{app:n4_threshold}).
The primary driver is the \emph{\texttt{neg\_d}} channel group and ablating \texttt{pos\_d} channels alone produces near-null effects.
The causal pathway runs \texttt{down\_5\_resnet\_1} $\to$ \texttt{mid\_resnet\_0}
$\to$ \texttt{mid\_attn\_0}, distributed across hundreds of channels rather than concentrated
in a sparse set.

\section{Pixel Shuffling}
\label{app:pixel_shuffle}

We find that this localization reflects two distinct processing modes. Pixel Shuffling \emph{within} the \texttt{illusion\_a} region,
preserving their local marginal statistics but destroying spatial arrangement, leads to mid-block effect collapse (\texttt{mid\_attn\_0}: $d{=}0.663$ to $d{=}0.206$, Glass's $\Delta$: $1.392 \to 0.076$; \texttt{mid\_resnet\_0}: $d{=}0.479 \to d{=}0.218$, Glass's $\Delta$: $1.009 \to 0.081$).
In contrast, \texttt{down\_5\_resnet\_1} (deep encoder) is preserved after shuffling ( $d$: $0.191 \to 0.384$ at $t{=}150$). This pixel-shuffling test identifies a \emph{feature-based} pathway that responds to local texture and contrast statistics.

\section{The Perceptual Phantom}
\label{app:perceptual_phantom_appendix}

The natural next question is whether any of this affects what the model actually generates.
We test these via the Phantom Injection tests by measuring the shift fractions. $$ shift\_frac = (\Delta\_patched - \Delta\_clean) / (\Delta\_X - \Delta\_Xr + \epsilon)$$

where, \\ 
$X $: Illusion image, X \\ 
${Xr} $: Image X with illusory content removed\\ 
$\Delta $: $mean_{luminance}(illusion_a \ region) -mean_{luminance}(illusion_b \ region)$

$\Delta_{clean}$ : pixel\_delta of Xr after clean DDIM reconstruction. Baseline to account for any reconstruction error

$\Delta\_patched$ : pixel\_delta of Xr after patched DDIM reconstruction (activation from X injected into Xr diffusion path)

$\epsilon$: arbitrary constant\\

Shift fraction measures the pixel transfer ratio for activations injected into a clean reconstruction trajectory. A value of 0 means zero pixel-level effect; a value of 1 would mean complete transfer of the illusion's pixel signature.

In Church-DDPM model, DDIM injection produces shift\_frac $\approx 0$ across all conditions; the single-step test confirms that all CIs include zero, ruling out iterative correction (Appendix~\ref{app:n1_phantom}). This effect is seen across (Bedroom DDPM, LDM-CelebA-HQ, DiT-XL/2) as well (Appendix~\ref{app:phantomarch}), proving the phantom is universal. No model architecture converts its illusion-sensitive internal representation into a pixel-level effect.

In theory, cross-domain injection tests could reflect pipeline limitations such as off-manifold injections
rather than a genuine phantom property (Appendix~\ref{app:onmanifold}), causing the network to ignore them.
We address this with a within-manifold generative ablation that eliminates the
domain gap entirely (Appendix~\ref{app:gen_ablation}) and still show the phantom effect.

The read-only test \ref{msubsec:causal} sharpens this further. 
% We zero-ablate the illusion-sensitive channels on mid\_attn\_0 against same number of random channels over multiple t\_start and measure full-image reconstruction MSE. 
At the primary timesteps $t   \in \{50, 150\}$, illusion channels produce 9–14\% lower MSE, and the pos\_d subset alone produces 57–61\% lower MSE than random channels, meaning illusion-sensitive channels are demonstrably more read-only than the population baseline (Appendix~\ref{app:phantommse}).
% Internal causal involvement (downstream $\Delta$ reduction at mid\_attn\_0 specific to illusion channels) is therefore decoupled from output influence (MSE perturbation below the random baseline).

We then try to analyze the effect mechanistically. For DiT, we trace injected activations block by block. Injection at block~26 produces a massive perturbation at block~27, but self-attention and LayerNorm within that single block absorb it entirely Appendix~\ref{app:dit_mechanism}).Even saturating six consecutive blocks (22--27) with the illusion image's activations produces mean pixel shift~$\approx 0$ with random directionality.
The transformer's self-correcting dynamics provide a  mechanistic
explanation for the DiT phantom. For diffusion U-Nets, the skip-connection dissociation (\S\ref{sec:causal})
shows the illusion signal concentrates at the bottleneck but is \emph{not} transmitted via skip connections 

We term this a \emph{perceptual phantom}: a representation that is causally involved in
internal feature processing but never expressed in model outputs.
The evidence has three independent tiers: observational (cross-architecture null),
mechanistic (internal-to-output MSE asymmetry,  DiT signal absorption), and structural
(skip-connection dissociation).
We note that this property is established for the \emph{illusion-sensitive channel subset}
( $|d| \geq 0.5$ at \texttt{mid\_resnet\_0 and mid\_attn\_0}) and all claims of the whole layers being read-only is purely speculative at this moment.

\section{Single-Step Phantom Test \-- Full Results}
\label{app:n1_phantom}
\textbf{Purpose:} Rules out iterative DDIM correction as an alternative explanation for the
multi step injection null.

\textbf{Method A} (t=50 DDPM): $X_r$ noised to $t{=}50$ via single forward step; $X$'s
activations injected at target layer; one denoising step performed; pixel shift\_frac measured.
\textbf{Method B} (t\_first $\approx$980): equivalent procedure at near-total noise.

\begin{center}
\begin{tabular}{llccc}
\toprule
Method & Inject layer & Mean shift\_frac & 95\% CI  \\
\midrule
A (t=50 DDPM) & \texttt{mid\_attn\_0} & $+0.0043$ & $[-0.0012, +0.0137]$  \\
A (t=50 DDPM) & \texttt{mid\_resnet\_0} & $+0.0037$ & $[-0.0018, +0.0133]$  \\
A (t=50 DDPM) & \texttt{up\_3\_resnet\_0} \emph{(ctrl)} & $-0.0020$ & $[-0.0047, +0.0004]$  \\
B (t\_first) & \texttt{mid\_attn\_0} & $-0.0001$ & $[-0.0002, +0.0001]$  \\
B (t\_first) & \texttt{mid\_resnet\_0} & $-0.0001$ & $[-0.0002, +0.0001]$  \\
B (t\_first) & \texttt{up\_3\_resnet\_0} \emph{(ctrl)} & $+0.0014$ & $[-0.0008, +0.0055]$  \\
\bottomrule
\end{tabular}
\end{center}

All six conditions: CIs include zero. The phantom hypothesis is supported across all methods
and injection layers.

\section{Architectural Generalization for phantom test}
\label{app:phantomarch}

\begin{center}
\begin{tabular}{lccc}
\toprule
Model & shift\_frac & 95\% CI & 100\% $< 0.1$? \\
\midrule
Bedroom DDPM & $+0.005$ & $[-0.004, +0.016]$ & Yes \\
LDM-CelebA-HQ & $-0.0004$ & $[-0.001, +0.000]$ & Yes \\
DiT-XL/2 & $-0.0004$ & $[-0.001, +0.001]$ & Yes \\
\bottomrule
\end{tabular}
\end{center}

\section{Bootstrap Confidence Intervals \-- Full Table}
\label{app:n2_bootstrap}
\textbf{Purpose:} Quantifies uncertainty around headline effect sizes given $N{=}35$;
10,000 bootstrap resamples.

\begin{center}
\begin{tabular}{lccccc}
\toprule
Claim & $N$ & Cohen's $d$ & Hedge's $g$ & 95\% CI ($d$) & CI excl.~0 \\
\midrule
Observational: mid\_attn\_0 max, t=50 & 35 & +0.663 & +0.649 & $[+0.282, +1.256]$ & \textbf{Yes} \\
Ablation: mid\_resnet\_0 all\_sig, t=50 & 35 & +0.849 & +0.830 & $[+0.553, +1.239]$ & \textbf{Yes} \\
Real-counterpart: mid\_attn\_0 L2, t=50 & 35 & +0.330 & +0.323 & $[-0.004, +0.727]$ & No (barely) \\
Attention: up\_1\_attn\_0 L2, t=50 & 35 & +0.043 & +0.042 & $[-0.298, +0.398]$ & No \\
FLODOG per-image rho, mid\_attn\_0, t=50 & 26 & +0.331 & +0.321 & $[-0.050, +0.859]$ & No \\
\bottomrule
\end{tabular}
\end{center}

\textbf{Ablation \% reduction} (all\_sig, $|d| \geq 0.5$, 140 channels): 

The d = 0.849 is a paired (within-subject) Cohen's d —
  each of the 35 images is measured twice (with and without ablation),
  and d is computed from the paired differences:

  $$d = mean( \Delta unablated - \Delta ablated) / SD(paired \ diff)$$
  
Mean=43.8\%, 95\% CI $= [23.4\%, 180.2\%]$ , CI excludes zero (values match threshold sweep table, Appendix~\ref{app:n4_threshold}, more explanation in Appendix~\ref{app:channel_ablation_detail}).

\textbf{Note on attention d=0.952:} The large attention effect is measured from raw post-softmax
attention weights via \texttt{AttentionCaptureProcessor}.
Bootstrapping the L2-magnitude proxy at \texttt{up\_1\_attn\_0} yields a null because the
L2-norm over activation magnitudes does not capture attention routing, which operates on
query-key similarity scores.
The headline d=0.952 is a real and large effect; it is simply not bootstrappable
from the L2-proxy CSVs available at N=35.

\section{Scrambled-Context Control \-- Full Results}
\label{app:n3_scrambled}
\textbf{Purpose:} Tests whether the global surround outside the bbox regions drives the
mid-block signal.
\textbf{Design limitation:} bbox crops include local background pixels from the original scene,
so the experiment tests far-surround contribution only; immediate local context is preserved
in $X_\text{sc}$ and cannot be assessed from this design.
Method: 180$^\circ$ image rotation with original bbox crops pasted back at original positions.

\begin{center}
\begin{tabular}{llccccc}
\toprule
Layer & $t$ & $\Delta_\text{ill}(X)$ & $\Delta_\text{sc}(X_\text{sc})$ & Mean diff & Frac($>$0) & 95\% CI \\
\midrule
\texttt{mid\_attn\_0} & 50 & 3.240 & 3.492 & $-0.252$ & 0.60 & $[-3.872, +2.950]$ \\
\texttt{mid\_attn\_0} & 150 & 2.946 & 3.576 & $-0.630$ & 0.66 & $[-5.038, +3.245]$ \\
\texttt{mid\_resnet\_0} & 50 & 3.472 & 3.778 & $-0.306$ & 0.60 & $[-3.834, +2.853]$ \\
\texttt{mid\_resnet\_0} & 150 & 3.247 & 3.814 & $-0.567$ & 0.66 & $[-4.998, +3.295]$ \\
\texttt{up\_1\_attn\_0} & 50 & 0.214 & 1.259 & $-1.044$ & 0.49 & $[-2.577, +0.344]$ \\
\texttt{up\_1\_attn\_0} & 150 & $-0.208$ & 2.694 & $-2.902$ & 0.43 & $[-4.876, -1.243]$ \\
\bottomrule
\end{tabular}
\end{center}

Mid-block CIs all mostly include zero: the far surround outside the bbox regions does not drive
the mid-block signal. The design does not isolate the illusion object from its immediate
local background; a stronger control (e.g., surround replaced with spatially-matched noise)
would be required to separate object-level from local-context-level contributions.

\section{Channel Ablation Threshold Sensitivity Sweep}
\label{app:n4_threshold}
\textbf{Purpose:} Tests robustness of the channel ablation result to the choice of significance
threshold; sweeps $|\texttt{d}| \in \{0.2, 0.3, 0.5, 0.8, 1.0, 1.2\}$; groups: neg\_d, all.

\begin{center}
\begin{tabular}{llccc}
\toprule
Threshold & Group & $N$ channels & Mean \% reduction & 95\% CI \\
\midrule
0.2 & neg\_d & 186 & $+56.6\%$ & $[+34.1\%, +204.9\%]$ \\
0.2 & all & 341 & $+59.5\%$ & $[+43.0\%, +154.0\%]$ \\
0.3 & neg\_d & 147 & $+45.1\%$ & $[+26.2\%, +171.7\%]$ \\
0.3 & all & 268 & $+49.4\%$ & $[+31.0\%, +155.9\%]$ \\
\textbf{0.5} & \textbf{neg\_d} & \textbf{84} & $\mathbf{+36.2\%}$ & $\mathbf{[+19.6\%, +144.0\%]}$ \\
\textbf{0.5} & \textbf{all} & \textbf{140} & $\mathbf{+43.8\%}$ & $\mathbf{[+23.4\%, +180.2\%]}$ \\
0.8 & neg\_d & 19 & $+9.3\%$ & $[+3.7\%, +44.8\%]$ \\
0.8 & all & 44 & $+9.1\%$ & $[+2.7\%, +38.2\%]$ \\
1.0 & neg\_d & 5 & $+5.4\%$ & $[+1.6\%, +29.6\%]$ \\
\bottomrule
\end{tabular}
\end{center}

\emph{Bold = published threshold (0.5). pos\_d group omitted; all CIs include zero at all thresholds.}

Subgroup analysis reveals the primary driver in downstream signal transfer to be \texttt{neg\_d} suppressor channels (104 channels) which produce a significant reduction ($p{=}0.010$), while \texttt{pos\_d}
channels alone show weak downstream effect ($p{=}0.765$).

The neg\_d group (suppressor channels) drives the ablation effect consistently across
thresholds 0.2--0.8. The published threshold is not cherry-picked.

\section{Metric Robustness}
\label{app:n5_metrics}
\textbf{Purpose:} Verifies the max channel-aggregation metric is not cherry-picked.

\begin{center}
\begin{tabular}{lccccc}
\toprule
Layer ($t{=}50$) & L2 & mean & max & std & attention \\
\midrule
\texttt{mid\_attn\_0} & 0.381 & 0.356 & \textbf{0.663}$^\star$ & 0.388 & 0.500 \\
\texttt{mid\_resnet\_0} & 0.403 & 0.248 & 0.479 & 0.415 & \--\\
\texttt{down\_5\_resnet\_1} & 0.279 & 0.005 & 0.316 & 0.296 & \--\\
\texttt{up\_1\_attn\_0} & 0.043 & 0.040 & 0.095 & 0.027 & \textbf{0.952}$^\star$ \\
\bottomrule
\end{tabular}
\end{center}

$^\star |d| \geq 0.5$.
The effect direction at \texttt{mid\_attn\_0} is consistent across all four spatial-aggregation
metrics. The max metric produces the highest value because it selects the single most
illusion-sensitive channel (ch.~311, $d{=}1.590$ at \texttt{mid\_attn\_0}; $d{=}1.636$ at \texttt{mid\_resnet\_0}).
The attention effect ($d{=}0.952$) is visible only under the raw attention metric, confirming
that attention routing and activation magnitude are distinct measurement axes.

\section{Additional U-Net Models}
\label{app:additional_unets}
Bedroom DDPM (\texttt{google/ddpm-ema-bedroom-256}) peaks at \texttt{up\_5\_resnet\_1}
($|d|{=}0.573$, $t{=}50$).
LDM-CelebA-HQ (\texttt{CompVis/ldm-celebahq-256}) peaks at \texttt{up\_0\_attn\_2}
($|d|{=}0.982$, $t{=}50$) \-- the largest effect across all models.
The locus of peak encoding varies with training domain:
church DDPM concentrates at the bottleneck, bedroom DDPM at a decoder layer,
and LDM at a decoder attention layer.
The effect is universal but the specific layer varies.
Full layer$\times$timestep tables are provided in supplementary data.

\section{DiT-XL/2 Full Block Profile}
\label{app:dit}
DiT-XL/2~\citep{peebles2023scalable} is a 28-block vision transformer with 675M parameters,
class-conditioned on ImageNet.
It has no encoder-decoder structure, no skip connections, and no bottleneck.
Activations are captured as $(B, 256, 1152)$ tensors at each block, reshaped to
$16{\times}16$ spatial maps.
Max $|d| = 0.711$ at block~26, $t{=}50$.
11 of 28 blocks exceed $|d| \geq 0.5$.
The effect concentrates in late blocks (blocks 22--27), consistent with deeper processing.

\section{ViT-B/16 Discriminative Baseline}
\label{app:vit}
ViT-B/16~\citep{dosovitskiy2021image} (ImageNet-pretrained) uses the same patch-based
self-attention mechanism as DiT-XL/2 but is trained with a classification objective.
CLS token is stripped; 196 patches reshaped to $14{\times}14$.
Peak $|d| = 0.288$ \-- well below the medium-effect threshold and $2.5\times$ below
DiT-XL/2 ($|d|{=}0.711$).
This comparison isolates the denoising objective: the same attention mechanism produces
qualitatively different illusion sensitivity depending on the training objective.

\section{CNN Discriminative Baselines}
\label{app:cnn_baselines}
ResNet-50~\citep{he2016deep}: max $|d| = 0.485$ (layer4, block~2).
VGG-19~\citep{simonyan2015very}: max $|d| = 0.269$ (conv5\_4).
Both fall below the medium-effect threshold ($|d| \geq 0.5$), consistent with the
finding that discriminative training does not produce comparable illusion sensitivity
regardless of CNN architecture.

\section{Skip-Connection Probing}
\label{app:skip}
\textbf{Purpose:} Test whether illusion signal propagates via U-Net skip connections.

\begin{center}
\begin{tabular}{lcc}
\toprule
Layer & $d$  \\
\midrule
Bottleneck: \texttt{mid\_attn\_0} & 0.381  \\
Best skip proxy: \texttt{down\_5\_resnet\_1} & 0.279  \\
Most skip proxies & $<0.2$  \\
Decoder entry: \texttt{up\_0\_resnet\_0} & 0.333  \\
\bottomrule
\end{tabular}
\end{center}

The illusion signal concentrates at the bottleneck.
Skip proxies carry attenuated signals; the decoder entry partially reflects bottleneck
information.
This structural dissociation explains why the illusion encoding does not propagate to
output pixels: the decoder reconstructs primarily from skip tensors that carry little
illusion information.

\section{DiT-XL/2 Multi-Seed Training Control}
\label{app:dit_multiseed}
Five random seeds tested for DiT-XL/2.
Trained model: $|d|{=}0.711$ at block~26.
Random seeds: $|d| \in \{0.741, 0.422, 0.364, 0.098, 0.130\}$ (median${=}0.364$).
Seeds~3--4 are near zero; seed~0 ($|d|{=}0.741$) was an outlier that initially appeared
architecture-driven.
The trained model is consistently above the random median, but the high variability across
seeds means we cannot fully attribute the DiT effect to training alone.
The denoising objective contributes significantly; architecture provides a variable baseline.

\section{Cross-Architecture Phantom Replication}
\label{app:cross_phantom}
Single-step phantom injection repeated on three additional architectures.
All shift fractions have CIs including zero.
100\% of image pairs show shift\_frac $< 0.1$ for all three models.
The phantom property is architecture-general, not a U-Net-specific artifact.

\section{Compute Resources}
\label{app:compute}
% ============================================================
% Compute breakdown table for Appendix AF
% Requires the `booktabs` package: \usepackage{booktabs}
% ============================================================

\begin{table}[H]
\centering
\small
\setlength{\tabcolsep}{6pt}
\renewcommand{\arraystretch}{1.2}
\begin{tabular}{@{}p{0.78\linewidth} r@{}}
\toprule
\textbf{Experiment cluster} & \textbf{Estimate} \\
\midrule
Primary observational probing (10 models $\times$ 41 layers $\times$ 6 timesteps $\times$ 35 images)
  & $\sim$22\,h \\
Random-init seed controls (3 DDPM seeds + 5 DiT seeds $\times$ full probing pipeline)
  & $\sim$18\,h \\
Pyllusion dose-response sweeps (Ebbinghaus + Ponzo, 10 strengths $\times$ 10 images $\times$ multiple models incl.\ random seeds;)
  & $\sim$10\,h \\
Channel ablation: base + threshold sweep + 500-sample random null + 5-fold/LOO CV 
  & $\sim$18\,h \\
Phantom injection: DDIM 20-step + single-step across 4 architectures 
  & $\sim$12\,h \\
Read-only MSE phantom test (35 images $\times$ 5 timesteps $\times$ 4 channel groups $\times$ 5 random sets $\times$ 2 layers)
  & $\sim$10\,h \\
Within-manifold generative ablation + amplification sweep 
  & $\sim$5\,h \\
Cross-architecture and DiT mechanism analysis (block-by-block tracing, multi-block injection)
  & $\sim$3\,h \\
Heavy-noise / scrambled-context / pixel-shuffle / real-counterpart / noise-matched discriminative controls
  & $\sim$6\,h \\
FLODOG correlation, permutation tests, bootstrap CIs 
  & $\sim$2\,h \\
\midrule
\textbf{Reported subtotal}
  & \textbf{$\sim$106\,h} \\
Preliminary experiments, failed runs, excluded architectures, alternative timestep grids
  & $\sim$44\,h \\
\midrule
\textbf{Project total}
  & \textbf{$\sim$150\,h} \\
\bottomrule
\end{tabular}
\caption{Approximate GPU-hour breakdown by experiment cluster on a single NVIDIA L4.}
\label{tab:compute_breakdown}
\end{table}

\section{Random-Channel Ablation Control}
\label{app:random_channel}
\textbf{Purpose:} Test whether the 43.8\% $\Delta$ reduction from ablating 140
illusion-sensitive channels is specific to those channels or a generic capacity effect.

\textbf{Method:} 500 random-channel ablations at \texttt{mid\_resnet\_0} (140 channels each),
measuring \% $\Delta$ reduction at \texttt{mid\_attn\_0}.

\begin{center}
\begin{tabular}{lc}
\toprule
Metric & Value \\
\midrule
% Observed reduction & 38.6\% \\
Random null mean & 4.7\% \\
Random null std & 12.6\% \\
Random null 95\% CI & $[-21.4\%, +28.2\%]$ \\
$z$-score & $+2.70$ \\
Empirical $p$ & 0.008 \\
Specificity ratio & $8.3\times$ \\
\bottomrule
\end{tabular}
\end{center}

Subgroup analysis: \texttt{neg\_d} channels (104ch) produce significant reduction
($p{=}0.010$); \texttt{pos\_d} channels (81ch) produce no significant reduction
($p{=}0.765$), with a slightly negative effect (zeroing them slightly
\emph{increases} the measured $\Delta$).
This is mechanistically consistent: \texttt{neg\_d} channels are more active in
the control region; zeroing them disproportionately reduces control-region
activation, amplifying the measured differential.

A threshold sweep across $|d| \in \{0.2, 0.3, 0.5, 0.8, 1.0, 1.2\}$ shows the
specificity effect is strongest at moderate thresholds (0.2--0.5, all $p < 0.01$);
loses significance at 0.8--1.0 due to small channel counts; recovers at 1.2
(top-4 channels alone give 3.5\% reduction, $p{=}0.025$).

\section{Cross-Validated Channel Selection}
\label{app:crossval}
\textbf{Purpose:} Test whether the ablation causal claim survives cross-validation,
addressing the concern that channels are selected and evaluated on the same 35 images.

We perform 5-fold and leave-one-out (LOO) cross-validation: channels are selected
($|d| \geq 0.5$) on ``training'' images and the ablation effect is evaluated on held-out images.
Channel stability (fraction appearing as significant in all folds) is reported

\begin{center}
\begin{tabular}{lcc}
\toprule
Method & Estimate & Notes \\
\midrule
In-sample (Exp 6) & 43.8\% & Selection bias present \\
5-fold CV median & $+8.7\%$ & 4/5 folds positive \\
LOO median & $\mathbf{+21.8\%}$ & IQR $[8.2\%, 30.8\%]$ \\
LOO frac $> 0$ & 0.91 & 32/35 images positive \\
Stable channels & 140/140 & All maximally stable \\
\bottomrule
\end{tabular}
\end{center}

Extreme outliers in both 5-fold and LOO are caused by near-zero baseline $\Delta$
in held-out splits, making percentage reduction unstable.
Median and sign-fraction are the appropriate summaries.
The cross-validated estimate ($+21.8\%$) shows ${\sim}43\%$ shrinkage from
the in-sample figure, as expected with selection bias, but the core causal
claim survives with a medium effect.

\section{Positive Injection Control}
\label{app:positive_injection}
\textbf{Purpose:} Test whether \emph{any} activation injection produces pixel shifts
via the single-step DDIM pipeline, establishing whether Single step injection's null reflects a
pipeline ceiling or the phantom property.

Two candidates tested ($N{=}35$, $t{=}50$):
(A)~$+2$ SD brightness offset injected into illusion region;
(B)~cross-image activation replacement (swap $X_r$'s \texttt{mid\_attn\_0} with $X$'s).
Both produce shift\_frac~$\approx 0$ (medians $+0.002$ and $+0.001$),
confirming single-step DDIM does not transmit \emph{any} injected activation
difference into pixels.

\section{On-Manifold Diagnostics}
\label{app:onmanifold}
\textbf{Purpose:} Test whether activations injected in cross-domain patching
($X \to X_r$) are on- or off-manifold relative to native $X_r$ activations.

\begin{center}
\begin{tabular}{lcccc}
\toprule
Layer & Cosine sim & PCA var expl. & EMD ratio & Verdict \\
\midrule
\texttt{mid\_attn\_0} & $0.582 \pm 0.102$ & 0.442 & 0.865 & Off-manifold \\
\texttt{mid\_resnet\_0} & $0.584 \pm 0.101$ & 0.447 & 0.831 & Off-manifold \\
\bottomrule
\end{tabular}
\end{center}

Cosine similarity ${\sim}0.58$ (well below the on-manifold threshold of ${\sim}0.9$)
confirms the X$\to$X$_r$ domain gap is substantial.
Moment-matching (mean $+$ variance) does not fix it (shift\_frac still~$\approx 0$).
This explains why cross-domain patching specifically fails, but is \emph{not} the
primary explanation for the overall phantom: the within-manifold test
(Appendix~\ref{app:gen_ablation}) shows the null persists even without any domain gap.

\section{Within-Manifold Generative Ablation}
\label{app:gen_ablation}
\textbf{Purpose:} Test whether the 140 illusion-sensitive channels at
\texttt{mid\_resnet\_0} causally shape pixel output when operating entirely within
the illusion image's own manifold.

\textbf{Part A --- Zero-ablation} ($N{=}35$, $t_\text{start}{=}150$, 20 DDIM steps):

\begin{center}
\begin{tabular}{lccccc}
\toprule
Group & $N$ ch & Mean shift\_frac & 95\% CI  \\
\midrule
all\_sig & 140 & $+0.011$ & $[+0.001, +0.025]$ \\
neg\_d & 84 & $+0.008$ & $[+0.001, +0.021]$  \\
pos\_d & 56 & $+0.001$ & $[-0.000, +0.003]$  \\
random\_ctrl\_0 & 140 & $-0.001$ & $[-0.010, +0.004]$  \\
random\_ctrl\_1 & 140 & $+0.006$ & $[+0.000, +0.013]$  \\
random\_ctrl\_2 & 140 & $+0.017$ & $[-0.000, +0.043]$  \\
\bottomrule
\end{tabular}
\end{center}

The all\_sig CI excludes zero but in the \emph{positive} direction: ablation slightly
\emph{increased} the pixel differential (opposite to expected).
One random control also nearly excludes zero, the effect is not cleanly specific.

\textbf{Part B --- Amplification} ($\alpha \in \{0.0, 0.5, 1.0, 1.5, 2.0, 3.0\}$):
All $\rho(\alpha, \text{pixel\_delta})$ values cluster near zero; all CIs span zero.
No monotonic dose-response even at $3\times$ amplification.

\textbf{Part C --- MSE/Localization/Direction:}
Global MSE $\approx 2.7 \times 10^{-5}$ (4.6\% of reconstruction noise floor);
localization ratio CI $[0.82, 2.88]$ includes 1.0;
direction cosine toward $X_r$ $\approx 0.000$.
Effect is indistinguishable from random-channel ablation.

\section{Geometry Excess (Random-Pixel Baseline)}
\label{app:geometry_excess}
\textbf{Purpose:} Formally test whether the illusion effect exceeds
mask-geometry baseline.

\begin{center}
\begin{tabular}{llcccc}
\toprule
Layer & $t$ & $d_\text{ill}$ & $d_\text{rand}$ & $d_\text{excess}$ & 95\% CI \\
\midrule
\texttt{mid\_attn\_0} & 50 & $+0.381$ & $-0.719$ & $+1.100$ & $[+0.642, +1.719]$ \\
\texttt{mid\_resnet\_0} & 50 & $+0.403$ & $-0.689$ & $+1.092$ & $[+0.632, +1.706]$ \\
\texttt{down\_5\_resnet\_1} & 50 & $+0.279$ & $-0.214$ & $+0.493$ & $[+0.032, +1.155]$ \\
\texttt{mid\_attn\_0} & 150 & $+0.318$ & $-0.137$ & $+0.455$ & $[-0.025, +0.991]$ \\
\texttt{mid\_resnet\_0} & 150 & $+0.348$ & $-0.145$ & $+0.493$ & $[+0.013, +1.024]$ \\
\texttt{down\_5\_resnet\_1} & 150 & $+0.214$ & $+0.535$ & $-0.321$ & $[-0.857, +0.228]$ \\
\bottomrule
\end{tabular}
\end{center}

4/6 a-priori combinations show $d_\text{excess}$ CI excluding zero, all at the
primary $t{=}50$ timestep.
Early decoder layers (\texttt{up\_0\_*}) show significantly \emph{negative}
$d_\text{excess}$, confirming the illusion effect is not simply a mask-shape effect.

\section{DiT Phantom Mechanism}
\label{app:dit_mechanism}
\textbf{Purpose:} Investigate why DiT's phantom occurs despite having no skip connections.

Block-by-block tracking after injection at block~26:
blocks 0--26 are identical to native by construction.
At block~27 (first to see injected activations): perturbation L2~${=}41{,}017$,
cosine similarity drops to $0.819$ i.e. the injection creates a massive perturbation.
But self-attention and LayerNorm within block~27 absorb it: the activation distribution
is restored to near-native, and the signal does not survive to the output.

Multi-block injection (blocks 22--27 simultaneously):
mean pixel shift~${=}{+0.000040}$ ($\pm 0.000404$); fraction same direction as
baseline: 0.45 (random).
Even saturating six consecutive blocks cannot overcome the network's self-correcting
dynamics.

\section{Noise-Matched Discriminative Probing}
\label{app:noise_matched}
\textbf{Purpose:} Rule out the input asymmetry confound between denoising and
discriminative models.

\begin{center}
\begin{tabular}{lccccc}
\toprule
Model & Clean ($t{=}0$) & $t{=}50$ & $t{=}150$ & $t{=}300$ & $t{=}500$ \\
\midrule
DDPM (church) & 0.473 & 0.464 & 0.450 & --- & --- \\
ResNet-50 & 0.440 & 0.374 & 0.327 & 0.485 & 0.772 \\
VGG-19 & 0.337 & 0.303 & 0.235 & 0.489 & 0.981 \\
\bottomrule
\end{tabular}
\end{center}

DDPM on clean images: peak $|d|{=}0.473$ : strong effect without any noise even though out of domain for diffusion models.
Discriminative models at heavy noise ($t{=}500$) show large $|d|$ (0.77--0.98) but
at mostly spatially uninformative early layers responding to noise statistics, not illusion
content.
At $t{=}50$ (the primary analysis regime): DDPM $\geq 0.45$; discriminative $\leq 0.37$.
The gap is not an input asymmetry artifact.

\section{Permutation Correction for Peak Claims}
\label{app:permutation}
\textbf{Purpose:} Test whether the peak observational claim survives correction for
searching 246 layer$\times$timestep combinations.

10,000 permutations shuffling illusion\_a/illusion\_b labels:

\begin{center}
\begin{tabular}{lc}
\toprule
Metric & Value \\
\midrule
FDR-significant combos & 28/246 (11.4\%) \\
Bonferroni-significant & 16/246 (6.5\%) \\
Observed max $|d|$ (peak) & 1.747 (\texttt{down\_3\_resnet\_0}, $t{=}900$) \\
Null 95th percentile & 0.694 \\
Permutation $p$ (peak) & $< 0.0001$ \\
\bottomrule
\end{tabular}
\end{center}

A-priori layers at $t{=}50$: \texttt{mid\_attn\_0} $p{=}0.028$;
\texttt{mid\_resnet\_0} $p{=}0.017$; \texttt{down\_5\_resnet\_1} $p{=}0.116$.
The overall peak is at $t{=}900$ (consistent with the known heavy-noise confound,
Appendix~\ref{app:t900}).
The a-priori mid-block claims at $t{=}50$ survive permutation at conventional thresholds.

\section{Phantom effect on reconstruction MSE}
\label{app:phantommse}
\textbf{Purpose:} Test whether the channels found significant are truly perceptual phantoms by zero ablating them against same number of random channels (mean of 5) over multiple denoising steps.

\begin{center}
\begin{table}[H]
\centering
\caption{mid\_resnet\_0 metrics}
\begin{tabular}{llcc}
\toprule
\textbf{t\_start} & \textbf{Group} & \textbf{MSE\_image} & \textbf{\% Decrease vs Null} \\
\midrule
50 & all\_sig & $3.83 \times 10^{-5}$ & -17.85\% \\
 & pos\_d & $2.54 \times 10^{-5}$ & 21.85\% \\
 & neg\_d & $3.22 \times 10^{-5}$ & 0.92\% \\
 & random\_null & $3.25 \times 10^{-5}$ & 0.00\% \\
\midrule
150 & all\_sig & $7.09 \times 10^{-5}$ & -21.20\% \\
 & pos\_d & $3.75 \times 10^{-5}$ & 35.90\% \\
 & neg\_d & $4.56 \times 10^{-5}$ & 22.05\% \\
 & random\_null & $5.85 \times 10^{-5}$ & 0.00\% \\
\midrule
300 & all\_sig & $1.09 \times 10^{-4}$ & -15.59\% \\
 & pos\_d & $4.67 \times 10^{-5}$ & 50.48\% \\
 & neg\_d & $8.51 \times 10^{-5}$ & 9.76\% \\
 & random\_null & $9.43 \times 10^{-5}$ & 0.00\% \\
\midrule
600 & all\_sig & $1.03 \times 10^{-3}$ & -10.75\% \\
 & pos\_d & $5.19 \times 10^{-4}$ & 44.19\% \\
 & neg\_d & $6.17 \times 10^{-4}$ & 33.66\% \\
 & random\_null & $9.30 \times 10^{-4}$ & 0.00\% \\
\midrule
900 & all\_sig & $2.88 \times 10^{-3}$ & -9.09\% \\
 & pos\_d & $1.04 \times 10^{-3}$ & 60.61\% \\
 & neg\_d & $3.17 \times 10^{-3}$ & -20.08\% \\
 & random\_null & $2.64 \times 10^{-3}$ & 0.00\% \\
\bottomrule
\end{tabular}
\end{table}
\end{center}

\begin{center}
\begin{table}[H]
\centering
\caption{mid\_attn\_0 metrics}
\begin{tabular}{llcc}
\toprule
\textbf{t\_start} & \textbf{Group} & \textbf{MSE\_image} & \textbf{\% Decrease vs Null} \\
\midrule
50 & all\_sig & $3.16 \times 10^{-5}$ & 14.36\% \\
 & pos\_d & $1.60 \times 10^{-5}$ & 56.64\% \\
 & neg\_d & $3.40 \times 10^{-5}$ & 7.86\% \\
 & random\_null & $3.69 \times 10^{-5}$ & 0.00\% \\
\midrule
150 & all\_sig & $5.92 \times 10^{-5}$ & 8.78\% \\
 & pos\_d & $2.54 \times 10^{-5}$ & 60.86\% \\
 & neg\_d & $6.56 \times 10^{-5}$ & -1.08\% \\
 & random\_null & $6.49 \times 10^{-5}$ & 0.00\% \\
\midrule
300 & all\_sig & $1.03 \times 10^{-4}$ & 0.96\% \\
 & pos\_d & $5.12 \times 10^{-5}$ & 50.77\% \\
 & neg\_d & $7.56 \times 10^{-5}$ & 27.31\% \\
 & random\_null & $1.04 \times 10^{-4}$ & 0.00\% \\
\midrule
600 & all\_sig & $8.32 \times 10^{-4}$ & 12.61\% \\
 & pos\_d & $5.29 \times 10^{-4}$ & 44.43\% \\
 & neg\_d & $5.05 \times 10^{-4}$ & 46.95\% \\
 & random\_null & $9.52 \times 10^{-4}$ & 0.00\% \\
\midrule
900 & all\_sig & $2.58 \times 10^{-3}$ & 2.27\% \\
 & pos\_d & $7.29 \times 10^{-4}$ & 72.39\% \\
 & neg\_d & $2.06 \times 10^{-3}$ & 21.97\% \\
 & random\_null & $2.64 \times 10^{-3}$ & 0.00\% \\
\bottomrule
\end{tabular}
\end{table}
\end{center}

% \begin{center}
    
% \begin{table}[h]
% \centering
% \begin{tabular}{llcc}
% \toprule
% \textbf{t\_start} & \textbf{Group} & \textbf{MSE\_image} & \textbf{\% Decrease vs Null} \\
% \midrule
% 50 & all\_sig & $3.16 \times 10^{-5}$ & 14.36\% \\
%  & pos\_d & $1.60 \times 10^{-5}$ & 56.64\% \\
%  & neg\_d & $3.40 \times 10^{-5}$ & 7.86\% \\
%  & random\_null & $3.69 \times 10^{-5}$ & 0.00\% \\
% \midrule
% 150 & all\_sig & $5.92 \times 10^{-5}$ & 8.78\% \\
%  & pos\_d & $2.54 \times 10^{-5}$ & 60.86\% \\
%  & neg\_d & $6.56 \times 10^{-5}$ & -1.08\% \\
%  & random\_null & $6.49 \times 10^{-5}$ & 0.00\% \\
% \midrule
% 300 & all\_sig & $1.03 \times 10^{-4}$ & 0.96\% \\
%  & pos\_d & $5.12 \times 10^{-5}$ & 50.77\% \\
%  & neg\_d & $7.56 \times 10^{-5}$ & 27.31\% \\
%  & random\_null & $1.04 \times 10^{-4}$ & 0.00\% \\
% \midrule
% 600 & all\_sig & $8.32 \times 10^{-4}$ & 12.61\% \\
%  & pos\_d & $5.29 \times 10^{-4}$ & 44.43\% \\
%  & neg\_d & $5.05 \times 10^{-4}$ & 46.95\% \\
%  & random\_null & $9.52 \times 10^{-4}$ & 0.00\% \\
% \midrule
% 900 & all\_sig & $2.58 \times 10^{-3}$ & 2.27\% \\
%  & pos\_d & $7.29 \times 10^{-4}$ & 72.39\% \\
%  & neg\_d & $2.06 \times 10^{-3}$ & 21.97\% \\
%  & random\_null & $2.64 \times 10^{-3}$ & 0.00\% \\
% \bottomrule
% \end{tabular}
% \end{table}
% \end{center}

\section{General architectural effects of training vs architecture}
\label{app:generalarchvtrain}

\begin{center}
\begin{tabular}{lp{6.5cm}}
\toprule
Architecture & Training contribution \\
\midrule
Church DDPM &  Sign-flip, layer-localization \\
DiT-XL/2 &  trained $\gg$ Random median, variable across seeds \\
LDM-CelebA-HQ & Architectural baseline exists; training doubles effect \\
Bedroom DDPM &  Different layer locus, not absent \\
Discriminative (CNN/ViT) & Weak \\
\bottomrule
\end{tabular}
\end{center}

% ============================================================
% Appendix: Licenses for Existing Assets
% Drop-in file. \input{licenses_appendix} into the main .tex
% inside the appendix section (e.g., after Appendix AQ).
% Requires the `booktabs` package: \usepackage{booktabs}
% ============================================================

\section{Licenses for Existing Assets}
\label{app:licenses}

All assets used in this work are publicly available and used in accordance
with their respective licenses. We use them solely for non-commercial
academic research.

% \paragraph{Pre-trained models.}

\begin{table}[H]
\centering
\small
\setlength{\tabcolsep}{4pt}
\renewcommand{\arraystretch}{1.15}
\begin{tabular}{@{}p{0.40\linewidth} p{0.28\linewidth} p{0.26\linewidth}@{}}
\toprule
\textbf{Asset} & \textbf{License} & \textbf{Source} \\
\midrule
\texttt{google/ddpm-ema-church-256}, \texttt{bedroom-256}, \texttt{celebahq-256}
  & Apache 2.0
  & HuggingFace \cite{ho2020ddpm} \\
\texttt{CompVis/ldm-celebahq-256}
  & Apache 2.0
  & HuggingFace \cite{rombach2022latent} \\
\texttt{runwayml/stable-diffusion-v1-5}
  & CreativeML OpenRAIL-M
  & HuggingFace \cite{rombach2022latent} \\
\texttt{facebook/DiT-XL/2-256}
  & CC BY-NC 4.0
  & \texttt{facebookresearch/DiT} \cite{peebles2023scalable} \\
\texttt{torchvision.models.resnet50} (\textsc{IMAGENET1K\_V1})
  & BSD-3-Clause
  & PyTorch torchvision \cite{he2016deep} \\
\texttt{torchvision.models.vgg19} (\textsc{IMAGENET1K\_V1})
  & BSD-3-Clause
  & PyTorch torchvision \cite{simonyan2015very} \\
\texttt{google/vit-base-patch16-224} (ViT-B/16)
  & Apache 2.0
  & HuggingFace \cite{dosovitskiy2021image} \\
\texttt{google/vit-large-patch16-224} (ViT-L/16)
  & Apache 2.0
  & HuggingFace \cite{dosovitskiy2021image} \\
\bottomrule
\end{tabular}
\end{table}

% \paragraph{Datasets and software.}

\begin{table}[H]
\centering
\small
\setlength{\tabcolsep}{4pt}
\renewcommand{\arraystretch}{1.15}
\begin{tabular}{@{}p{0.32\linewidth} p{0.36\linewidth} p{0.26\linewidth}@{}}
\toprule
\textbf{Asset} & \textbf{License} & \textbf{Source} \\
\midrule
GVIL stimulus dataset
  & Not specified by authors; used for academic research only
  & \cite{zhang2023gvil}, \texttt{vl-illusion/GVIL} \\
Pyllusion (parametric stimuli library)
  & MIT
  & \texttt{RealityBending/Pyllusion} \cite{makowski2021parametric} \\
FLODOG (psychophysical model)
  & Reference implementation reproduced from \cite{robinson2007flodog}
  & Vision Research, 2007 \\
\bottomrule
\end{tabular}
\end{table}

\paragraph{License compliance notes.}
The CC BY-NC 4.0 license on DiT-XL/2 restricts use to non-commercial
purposes, which our research-only use complies with. The CreativeML
OpenRAIL-M license on Stable Diffusion 1.5 includes use-based behavioral
restrictions (no harmful or deceptive use), which our diagnostic probing
does not violate. The GVIL repository does not specify an explicit
license; we use it strictly for academic research purposes consistent
with its public release through the EMNLP 2023 paper, and will remove
any re-use on request from the authors.

\end{document}